\def\useplain{1}
\ifx \useplain\undefined
	\documentclass[preprint, nonblindrev]{style/informs3aa}
	\usepackage{macros/informs_set}
	
\TITLE{Query Complexity of Bayesian Private  Learning}
\RUNTITLE{Query Complexity of Bayesian Private  Learning}

\RUNAUTHOR{Xu}

\ARTICLEAUTHORS{%
\AUTHOR{Kuang Xu}
\AFF{Graduate School of Business\\
Stanford University\\  
\EMAIL{kuangxu@stanford.edu}}  
} 

\else
	\documentclass[11pt]{article}
	\usepackage[margin=1.5in]{geometry}
	\usepackage{macros/plain_macro}
	\title{Query Complexity of Bayesian Private  Learning}

\author{
Kuang Xu\\
 Graduate School of Business\\
Stanford University \\
 \texttt{kuangxu@stanford.edu}  
} 

\date{}

\fi

\usepackage{macros/kuang_macro}

\begin{document}

\ifx \useplain\undefined
	
\TITLE{Query Complexity of Bayesian Private  Learning}
\RUNTITLE{Query Complexity of Bayesian Private  Learning}

\RUNAUTHOR{Xu}

\ARTICLEAUTHORS{%
\AUTHOR{Kuang Xu}
\AFF{Graduate School of Business\\
Stanford University\\  
\EMAIL{kuangxu@stanford.edu}}  
} 

\else
	\title{Query Complexity of Bayesian Private  Learning}

\author{
Kuang Xu\\
 Graduate School of Business\\
Stanford University \\
 \texttt{kuangxu@stanford.edu}  
} 

\date{}

\fi

\ifx \useplain\undefined
\else
\maketitle
\fi

\def\dl#1{{}}
\def\xk#1{{\color{blue} #1}}
\def\comm#1{{\color{red} (KX: #1)}}

\def\abs_txt{We study the query complexity of Bayesian Private Learning: a learner wishes to locate a random target within an interval by submitting queries, in the presence of an adversary who observes all of her queries but not the responses. How many queries are necessary and sufficient in order for the learner to accurately estimate the target, while simultaneously concealing the target from the adversary?

Our main result is a query complexity lower bound that is tight up to the first order. We show that if the learner wants to estimate the target within an error of $\veps$, while ensuring that no adversary estimator can achieve a constant additive error with probability greater than $1/L$, then the query complexity is on the order of $L\log(1/\veps)$, as $\veps \to 0$. Our result demonstrates that increased privacy, as captured by $L$, comes at the expense of a {multiplicative} increase in query complexity. 
 
Our proof  method builds on Fano's inequality and a family of proportional-sampling estimators. As an illustration of the method's wider applicability, we generalize the complexity lower bound to settings involving high-dimensional linear query learning and  partial adversary observation.\footnote{This version: October 2019. A preliminary version appeared in the proceedings of the \emph{Conference on Neural Information Processing Systems (NeurIPS)}, December 2018 \citep{xu2018query}.}\\
\emph{Keywords}: private active learning, goal-oriented privacy, action-information proximity, stochastic systems.
}

\ifx \useplain\undefined
\ABSTRACT{\abs_txt}
\else
\begin{abstract}
\abs_txt
\end{abstract}
\fi

\ifx \useplain\undefined
\maketitle
\fi

\section{Introduction}
\label{sec:intro}

How to learn, while ensuring that a spying adversary does not learn?   Enabled by rapid advancements in the Internet, surveillance technologies and machine learning,  companies or governments alike have become increasingly capable of monitoring the behavior of individuals, consumers and competitors, and use such data for inference and prediction. Motivated by these developments, the present paper investigates the extent to which it is possible for a learner to protect her knowledge from an adversary who observes, completely or partially, her actions. Furthermore, we are interested in how much additional effort is required for the learner to retain such privacy, and conversely, from the adversary's perspective, what kinds of statistical inference algorithms are most effective against a potentially privacy-conscious subject. 

Concretely, we approach these questions by studying the query complexity of Bayesian Private Learning, a framework proposed by \cite{xu2017private} and \cite{tsixuxu2017} to investigate the privacy-efficiency trade-off in active learning. Our main result is a tight lower bound on query complexity, showing that there will be a price to pay for the learner in exchange for improved privacy, whose magnitude scales {multiplicatively} with respect to the level of privacy desired. In addition, we provide a family of inference algorithms for the adversary, based on proportional sampling, which is provably effective in estimating the target against any learner who does not employ a large number of queries.  

Among our chief motivations are applications that involve protecting the privacy of consumers or firms against increasingly powerful surveillance and data analysis technologies; these applications are discussed in Section \ref{sec:applications}.

\subsection{The Model: Bayesian Private  Learning}
We begin by describing the Bayesian Private  Learning model formulated by  \cite{xu2017private} and \cite{tsixuxu2017}. A \emph{learner}  is trying to accurately identify the location of a random \emph{target}, $X^*$,  up to some constant additive error, $\veps$, where  $X^*$ is uniformly distributed in the unit interval, $[0, 1)$. The learner gathers information about $X^*$ by submitting $n$ \emph{queries}
\begin{equation}
 (Q_1,\ldots, Q_n) \in [0, 1)^n,  
 \end{equation} 
for some $n\in \N$. For each query, $Q_i$, she receives a binary \emph{response}, indicating the target's  location relative to the query: 
\begin{equation}
R_i = \mathbb{I}(X^*\leq Q_i), \quad i = 1, 2,\ldots, n, 
\label{eq:respDef}
\end{equation}
where $\mathbb{I}(\cdot)$ denotes the indicator function. Viewed from the angle of optimization, the target can be thought of as the minimum of a convex function with unknown parameters, and the responses will correspond to the signs of the  function's gradients at the queried points.

The learner submits the queries in a sequential manner, and subsequent queries may depend on previous responses. Once all $n$ queries are submitted, the learner will produce an estimator for the target. The learner's behavior is formally captured by a learner strategy, defined as follows. 

\begin{definition}[Learner Strategy]
\label{def:learnerContinuous}
 Fix $n \in \N$. Let $Y$ be a uniform random variable over $[0, 1)$, independent from other parts of the system; $Y$ will be referred to as the random seed.   A \emph{learner strategy}, $\phi = (\phi^q, \phi^l)$, consists of two components: 
\begin{enumerate}
\item Querying mechanism: $ \phi^q= (\phi^q_1,\ldots, \phi^q_n)$, is a sequence of deterministic functions, where $\phi^q_i: [0, 1)^{i-1}\times [0, 1) \to [0, 1)$ takes as input past responses and the random seed, $Y$, and generates the next query, i.e.,\footnote{Note that the query $Q_i$ does not explicitly depend on previous queries, $\{Q_1,\ldots, Q_{i-1}\}$, but only their responses. This is without the loss of generality, since for a given value of $Y$ it is easy to see that $\{Q_1,\ldots, Q_{i-1}\}$ can be reconstructed once we know their responses and the functions $\phi_1^q,\ldots, \phi_n^q$.} 
\begin{equation}
Q_i = \phi^q_i(R^{i-1}, Y), \quad i = 1, \ldots, n, 
\end{equation}
where $R^i$ denotes the responses from the first $i$ queries: $R^i= (R_1, \ldots, R_i)$, and $R^0 \bydef \emptyset$. 
\item Estimator: $\phi^l: [0, 1)^{n}\times [0, 1) \to [0, 1)$ is a deterministic function that maps all responses, $R^{n}$, and $Y$ to a point in the unit interval that serves as a ``guess'' for $X^*$: 
\begin{equation}
\what{X} = \phi^l(R^n, Y). 
\end{equation}
 $\what{X}$ will be referred to as the learner estimator.
\end{enumerate}

We will use  $\Phi_n$ to denote the family of learner strategies that submit $n$ queries. 
\end{definition}

The first objective of the learner is to accurately estimate the target, as is formalized in the following definition. 
\begin{definition}[$\veps$-Accuracy] 
\label{def:eps-acc-cont}
Fix $\veps \in (0,1)$. A learner strategy, $\phi$, is $\veps$-accurate, if its estimator approximates the target within an absolutely error of $\veps/2$ almost surely, i.e.,
\begin{equation}
\pb\lt(\lt| \what{X} -  X^*\rt| \leq \veps/2 \rt) = 1, 
\end{equation}
where the probability is measured with respect to the randomness in the target, $X^*$, and the random seed, $Y$. 
\end{definition}

We now introduce the notion of \emph{privacy}: in addition to estimating $X^*$, the learner would like to simultaneously conceal $X^*$ from an eavesdropping adversary. Specifically, there is an adversary who knows the learner's query strategy, and observes {all} of the queries but not the responses. The adversary then uses the query locations to generate her own \emph{adversary estimator} for $X^*$, denoted by $\what{X}^a$, which depends on the queries, $(Q_1,\ldots, Q_n)$, and any internal, idiosyncratic randomness. 

With the adversary's presence in mind, we  define the notion of a private learner strategy. 

\begin{definition}[$(\delta,L)$-Privacy]
\label{def:dLprivacy}
Fix $\delta \in (0,1)$ and $L \in \N$. A learner strategy, $\phi$, is $(\delta,L)$-private if, for \emph{any} adversary estimator, $\what{X}^a$, 
\begin{equation}
\pb(|\what{X}^a - X^*| \leq \delta/2) \leq 1/L, 
\end{equation}
where the probability is measured with respect to the randomness in the target, $X^*$, and any randomness employed by the learner strategy and the adversary estimator.\footnote{This definition of privacy is reminiscent of the error metric used in  Probably Approximately Correct (PAC) learning \citep{valiant1984theory}, if we view the adversary as trying to learn a (trivial) constant function to within an $L_1$ error of $\delta/2$ with a probability great than $1/L$. }
\end{definition}

In particular, if a learner employs a $(\delta,L)$-private strategy, then no adversary estimator can be close to the target within an absolute error of $\delta/2$  with a probability great than $1/L$. Therefore, for any fixed  $\delta$, the parameter $L$ can be interpreted as the \emph{level of desired privacy}.  

We are now ready to define the main quantity of interest in this paper: query complexity. 
\begin{definition} Fix $\veps$ and $\delta$ in $ [0,\, 1]$, and $L \in \N$. The \emph{query complexity}, $N(\veps, \delta, L)$, is the least number of queries needed for an $\veps$-accurate learner strategy to be $(\delta, L)$-private: 
\begin{equation}
N(\veps, \delta, L) \bydef \min\{n: \, \Phi_n \mbox{ contains a  strategy that is both $\veps$-accurate and $(\delta, L)$-private}\}. \nln
\end{equation}
\end{definition}

\begin{remark}[Connections to Active Learning] 
\label{remk:activelearn}
Cast in the terminology of active learning, the problem facing the learner is equivalent to that of learning a   {threshold function}, $f_{X^*}(\cdot)$, with the threshold $X^*$: $f_{X^*}(x) = \mathbb{I}( X^* \leq x)$, $x\in [0,1)$. The response $R_i$ is simply the value of the function evaluated at $Q_i$:  $f_{X^*}(Q_i)$. A learner strategy is $\veps$-accurate if the leaner is able to produce a threshold function $\hat{f}$ such that $\|\hat f - f_{X^*}\|_1 \bydef \int_{x\in (0,1]} |\hat f (x)  - f_{X^*}(x)| dx \leq \veps/2.$
\end{remark}

\subsection{Notation and Convention}
We will use the asymptotic notation $f(x) \sim g(x)$ to mean that $f$ is on the order of $g$: $f(x)/g(x) \to 1$ as $x$ approaches a certain limit. All logarithmic functions used in this paper will be with base $2$. To avoid excessive use of floors and ceilings, we will assume the values of $\veps$, $\delta$ and $1/L$ are integral powers of $1/2$, so that their corresponding logarithmic expressions always assume an integral value. When referring to an interval that belongs to a partition of $[0,1)$, we will use the term ``sub-interval'' to distinguish it from the unit interval itself; the same is true with the term ``sub-cube'' in higher dimensions. We use $x\vee y$ and $x \wedge y$ as a short-hand for $\max\{x, y\}$ and $\min \{ x, y\}$, respectively. 

\section{Main Result}

The main objective of the paper is to understand how $N(\veps, \delta, L)$ varies as a function of the input parameters, $\veps$, $\delta$ and $L$. Our result will focus on the regime of parameters where 
\begin{equation}
0<\veps< \delta/4 \, \mbox{ and } \, \delta < 1/L. 
\end{equation}
Having $\veps< \frac{1}{4}\delta$ corresponds to a setting where the {learner} would like to identify the target with high accuracy, while the adversary is aiming for a coarser estimate; the specific constant $\frac{1}{4}$ is likely an artifact of our analysis and could potentially be improved to being closer to $1$. Note that the regime where $\veps > \delta$ is arguably much less interesting, because  it is not natural to expect the adversary, who is not engaged in the querying process, to have a higher accuracy requirement than the {learner}. The requirement that $\delta < 1/L$ stems 
{from the 
following argument. If $\delta > 1/L$, then the adversary can simply draw a point uniformly at random in $[0, 1)$ and be guaranteed that the target will be within $\delta/2$ with a probability greater than $1/L$.   Thus, the privacy constraint is automatically  violated, and no private {learner} strategy exists. 
To obtain a nontrivial problem, we therefore need only to consider the case where $\delta < 1/L$. 
}

The following theorem is our main result. The upper bound has appeared in \citet{xu2017private} and  \citet{tsixuxu2017} and is included for completeness; the lower bound is the contribution of the present paper.
 
\begin{theorem}[Query Complexity of Bayesian Private Learning] 
\label{thm:main}
Fix  $\veps$ and $\delta$ in  $ (0, 1)$ and $L \in \N$, such that $\veps < \delta/4$ and $\delta < 1/L$. The following is true.
\begin{enumerate}
\item Upper bound: 
\begin{equation}
 N(\veps, \delta, L) \leq   L\log(1/\veps) - L(\log L -1)-1. 
\end{equation}
\item Lower bound: 
\begin{equation}
N(\veps, \delta, L)\geq  L\log(1/\veps) - L\log(2/\delta)- 3L\log\log(\delta/\veps).
\end{equation}
\end{enumerate}
\end{theorem}
Both the upper and lower bounds in Theorem \ref{thm:main} are constructive, in the sense that we will describe a concrete learner strategy that achieves the upper bound, and an adversary estimator that forces any learner strategy to employ at least as many queries as that prescribed by the lower bound. 

If we apply Theorem \ref{thm:main} in the regime where $\delta$ and $L$ stay fixed, while the learner's error tolerance, $\veps$, tends to zero, we obtain the following corollary in which the upper and lower bounds on query complexity coincide. 

\begin{corollary} 
\label{cor:main}
Fix $\delta\in (0, 1)$ and $L\in \N$, such that $ \delta < 1/L$. Then, 
\begin{equation}
N(\veps, \delta, L)  \sim  L\log(1/\veps), \quad \mbox{as $\veps \to 0$}. 
\end{equation}
\end{corollary}

Note that the special case of $L=1$ corresponds to  when the learner is not privacy-constrained and aims to solely minimize the number of queries. Theorem \ref{thm:main} and Corollary \ref{cor:main} thus demonstrate that there is a hefty price to pay in exchange for privacy, as the query complexity depends {multiplicatively} on the level of privacy,  $L$.  

Deriving the query complexity formulae in Theorem \ref{thm:main} is not the only objective of our inquiry. As will become clearer,  the proof of Theorem \ref{thm:main} will lead us to discovering the surprising efficacy of certain, seemingly naive, adversary estimators based on proportional sampling, and through them, we will obtain new insights into the problem's strategic dynamics. Consequently, we will be able to better answer questions such as: what are the key features that make it hard to conceal (learned) information in a sequential learning problem? How should the adversary take advantage of these features when designing inference algorithms? A key concept along this direction, that of \emph{information-action locality}, will be further explored   in Section \ref{sec:AI_local}. 

\begin{remark}[Noisy vs.~Noiseless Responses] 
Our model assumes that the responses, $R_i$, are {exact} (Eq.~\eqref{eq:respDef}), in contrast to some of the noisy response models in the literature \citep[e.g.,][]{rivest1980coping, ben2008bayesian, waeber2013bisection}, where, for instance, the true responses are flipped with a positive probability. We observe that the query complexity  lower bound  in Theorem \ref{thm:main} automatically applies  to the noisy setting (because  a learner's strategy can simulate noisy query responses  by artificially perturbing the true responses), while the upper bound doesn't (because it would require a different set of private learner strategies). We focus on the noiseless model because it is an important baseline that allows us to streamline the analysis and brings to the fore key insights. That being said, generalizing our results to a noisy query model can be an interesting and practically relevant direction of future research; it is discussed further in Section \ref{sec:conclusion}. 
\end{remark}

\section{Motivating Applications}
\label{sec:applications}

While our model is stylized, it is designed to capture fundamental privacy vs.~complexity tradeoffs inherent in applications of sequential learning. We examine below two such  motivating applications: 

{\bf Example 1: Protecting consumer privacy}. Consider a consumer (the learner) who is browsing an online retailer site in search of an item with an ``ideal'' one-dimensional feature value (the target), such as its size, brightness, or color tone. While the consumer does not directly know what the ideal value is, when presented with an item, she is able to articulate the ``impression'' as to whether the current item's feature value  is too large or to small. Guided by these impressions, the consumer browses different items in a sequential manner to eventually narrow down  the ideal item. During this process, the online retailer observes all of the items viewed by the consumer along with their feature values, but does not know the  impressions perceived by the consumer. Can the consumer conduct the search in such a way that does not reveal to the retailer the ideal item that she ultimately intends to purchase? (One can imagine that such information would put the consumer in a disadvantageous position for various reasons.) How many additional items should the consumer browse, and in what manner, should she wish to achieve successful obfuscation against the retailer's inference algorithms? 

{\bf Example 2: Privacy-aware price learning} (\citet{tsixuxu2017}). Consider a firm (the learner) is in the process of launching a new product, and would like to price the product in a way that maximizes total profit. The profit, $f(p)$, is a concave function of the price, $p$, and since the parameters of $f$ are unknown, the firm believes the profit maximizing price, $p^*=\argmax_{p} f(p)$, to be  uniformly distributed in a certain interval, $\calI$. To identify $p^*$ (corresponding to the target), the firm proceeds to test the sensitivity of the profit function at a sequence of price points $(Q_1,\ldots, Q_n) \in \calI^n$ using costly surveys or experiments; the test prices correspond to the queries and the resulting sensitivities the responses. Because the testing prices may be easily obtained or even public, the firm is concerned that a competitor (the adversary) who observes the testing prices will be able to predict $p^*$ and price their competing products accordingly, which would be  detrimental to the firm. The firm would thus like to minimize the number of testing prices, while ensuring that $p^*$ remains unpredictable for the competitors.

Finally, let us be reminded that Bayesian Private Learning  is a more general model that contains, as a special case ($L=1$), the classical problem of sequential learning with binary feedback. The latter has a wide range of applications in statistics \citep{robbins1951stochastic}, information theory \citep{horstein1963sequential} and  optimization \citep{waeber2013bisection}, and as a more general model Bayesian Private Learning inherits these applications as well. 


\section{Related Literature}

Bayesian Private Learning is a variant of the so-called Private Sequential Learning problem. Both models were formulated in \citet{xu2017private} and \citet{tsixuxu2017}, and the main distinction between the two is that the target is drawn \emph{randomly} in Bayesian Private Learning, while it is chosen in a \emph{worst-case} fashion (against the adversary) in the original Private Sequential Learning model.  \citet{xu2017private} and \citet{tsixuxu2017} establish matching upper and lower bounds on query complexity for Private Sequential Learning. They also propose the Replicated Bisection algorithm as a learner strategy for the Bayesian variant, but without a matching query complexity lower bound. The present paper closes this gap. 

The two formulations indeed differ in crucial ways. The worst-case assumption in the original model imposes a more stringent criterion for the adversary: she would have to ensure a certain probability of correct estimation for \emph{any} realization of the target (more risk-averse). In contrast, the adversary in the Bayesian version only has to do so {on average} (more risk-neutral). This distinction further leads to very different query complexities:  the query complexity in the worst-case formulation was shown to be around $\log(\delta/\veps)+2L$ \citep{xu2017private, tsixuxu2017}, whereas we show that the query complexity in the Bayesian setting is approximately $L\log(\delta/\veps)$ (Theorem \ref{thm:main}). That is, the privacy requirement demands  significantly greater efforts from the learner in the Bayesian version, from being additive to multiplicative in the level of privacy, $L$. Finally, the proof techniques for establishing lower bounds in the two settings also diverge: the arguments employed by  \citet{xu2017private} and \citet{tsixuxu2017} are combinatorial in nature, whereas our proof relies on information-theoretic tools.  

At a higher level, our work is connected to a growing body of literature  on privacy-preserving  mechanisms,  in computer science \citep{dwork2014algorithmic,lindell2009secure, fanti2015spy}, operations research \citep{cummings2016empirical,tsixu2018}, and statistical learning theory \citep{chaudhuri2011differentially,jain2012differentially,wainwright2012privacy}.
Beyond the more obvious divergence in models and applications, we highlight below some conceptual differences between Bayesian Private Learning and the extant literature.

1.  \emph{Goal-Oriented vs.~Universal}: this is the most  distinguishing feature of our privacy framework. Our notion of privacy is \emph{goal-oriented}, defined with respect to the adversary's (in)ability to perform a {specific} statistical inference task. 
Notably, this is in contrast to the (much more stringent) {universal} privacy criteria, among which Differential Privacy \citep[e.g.,][]{dwork2014algorithmic} is a well-known paradigm, where the output  distribution of a mechanism is supposed to be insensitive with respect to any perturbation in the input, and hence preventing an adversary from performing \emph{any} inference task.  A key consequence of the context-dependent formulation is that  the decision strategy can be tailored  to the adversary's inference task, and hence more {efficient}; in contrast, universal private requires the system designer to defend against a wider range of possible adversaries, and thus restricts  the designer to using only conservative and inefficient strategies.  The difference between the context-dependent and universal privacy is further explored in Appendix \ref{sec:BPSnotDP}: we show that an $(\delta,L)$-private  Replicated Bisection strategy is \emph{never} differentially private, thus demonstrating that the latter is a strictly more restricting privacy criterion. 

2. \emph{Concealing the Unknown vs.~the Known}: the object to be concealed in our model is initially unknown even to the decision maker herself, and must be {actively learned}. This is in contrast to privacy models where  the decision maker knows in advance the information to be concealed: examples of such information include the rumor source in the anonymous rumor spreading problem of \citet{fanti2015spy}, or the goal vertex in the Goal Prediction game  of \citet{tsixu2018}. 

3. \emph{Sequential vs.~One-shot}: we focus on {sequential} and dynamic learning, as opposed to one-shot or static problems \citep[e.g.,][]{gupta2012iterative}. 

 4. \emph{Decision-centric vs.~Data-centric}: we focus on the behavior and actions of a privacy-aware {decision maker}, as opposed to anonymizing a \emph{data set}  \citep{gasarch2004survey,dwork2008differential,gupta2012iterative}. The obfuscation measures in a decision problem can be significantly more limited than those in privatizing a data set: one is only able to modify {which} actions are chosen and not how they are observed, whereas a data release algorithm could inject richer noises or fudge data entries. This is an inherent byproduct of the  fact that the decision maker often has to use the actions as a means to acquire information or achieve a goal, and sometimes reveals the actions directly to the adversary (e.g., a search engine or data provider), rendering injecting arbitrary noise virtually impossible. 

On the methodological front, our proof uses Fano's inequality, an essential tool for deriving lower bounds in statistics, information theory, and active learning \citep{cover2012elements}. The proportional sampling estimators that we analyze are reminiscent of the reward-matching policy studied by \citet{xu2016reinforcement} and,  more broadly, the so-called Luce's rule in reinforcement learning \citep{luce1959individual}, where actions are chosen with probabilities proportional to the amount of associated rewards; these policies are known to perform well in repeated games \citep{erev1998predicting}. However, to the best of our knowledge, both proportional-sampling estimators and the use of Fano's inequality have received relatively little attention in the context of private sequential learning.  

Finally, we remark that recently, building on the conference version of the present manuscript \citep{xu2018query},  the authors of \cite{xu2019optimal} were able to use more intricate analysis and a variant of the proportional sampling estimator proposed in this paper, which they termed the truncated proportional sampling estimator, to established a pair of improved  query complexity upper and lower bounds  that are tight up to an additive factor of $\mathcal{O}(L)$.

\section{Proof Overview}
\label{sec:proofOverview}

The next two sections are devoted to the proof of Theorem \ref{thm:main}. We first give an overview of the main ideas. Let us begin by considering the  special case of $L=1$, where learner is solely interested in finding the target, $X^*$, and not at all concerned with concealing it from the adversary. Here, the problem reduces to the classical setting, where it is well-known that the bisection strategy achieves the optimal query complexity \citep{waeber2013bisection}. The bisection strategy recursively queries the mid-point of the interval which the learner knows to contain $X^*$. For instance, the learner would set $Q_1=1/2$, and if the response $R_1 = 0$, then she will know that $X^*$ lies in the interval $[0,1/2]$, and set $Q_2$ to $1/4$; otherwise, $Q_2$ will be set to $3/4$. This process repeats for $n$ steps.  Because the size of the smallest interval known to contain $X^*$ is halved with each additional query, this yields the query complexity
\begin{equation}
 N(\veps, \delta, 1) = \log(1/\veps), \quad \veps\in (0,1). 
 \label{eq:nopriv}
 \end{equation} 

Unfortunately, once the level of privacy $L$ increases above $1$, the bisection strategy is almost \emph{never} private: it is easy to verify that if the adversary sets $\what{X}^a$ to be  the learner's last query, $Q_n$, then the target is sure to be within a distance of at most $\veps$. That is, the bisection strategy is not $(\delta, L)$-private for any $L>1$, whenever $\veps<\delta/2$. This is hardly surprising: in the quest for efficiency, the bisection strategy submits queries that become progressively closer to the target, thus rending its location obvious to the adversary. 

Building on the bisection strategy, we arrive at a natural compromise: instead of a single bisection search over the entire unit interval, we could create $L$ identical copies of a bisection search across  $L$ disjoint sub-intervals of $[0,1)$ that are chosen ahead of time, in a manner that makes it impossible to distinguish which search is truly looking for the target. This is the main idea behind the Replicated Bisection strategy, first proposed and analyzed in \citet{tsixuxu2017}. We examine this strategy in Section \ref{sec:upperbound}, which will yield the query-complexity upper bound, on the order of $L\log({1}/{\veps})$.  

Proving the lower bound turns out to be more challenging. To show that the query complexity is at least, say, $n$, we will have to demonstrate that none of the learner strategies using $n-1$ queries, $\Phi_{n-1}$, can be simultaneously private and accurate. Because the sufficient statistic for the adversary to perform estimation is the posterior distribution of the target given the observed queries, a frontal assault on the problem would require that we characterize the resulting target posterior distribution for {all} strategies, a daunting task given the richness of $\Phi_{n-1}$, which grows rapidly as $n$ increases. 

Our proof will  take an indirect approach. The key idea is that, instead of allowing the adversary to use the entire posterior distribution of the target, we may restrict her to a seemingly much weaker class of \emph{proportional-sampling estimators}, where the estimator $\what{X}^a$ is sampled from a distribution proportional to the {empirical density} of the queries. A proportional-sampling estimator would, for instance,  completely ignore the order in which the queries are submitted, which may contain useful information about the target. We will show that, perhaps surprisingly, the proportional-estimators are so powerful that they leave the learner no option but to use a large number of samples. This forms the core of the lower bound argument. 

Studying the proportional-sampling estimators has additional benefits. From a practical perspective, they are constructive estimators that are extremely easy to implement and yet guaranteed to provide good estimation accuracy against \emph{any} learner strategy that uses few queries. More importantly, their structure avails us with deeper insights into a fundamental dilemma the learner faces: in order to acquire a sufficient amount of information to locate the target accurately, a significant portion of the learner's actions (queries) must be {spatially close} to the said target. It is precisely this  \emph{action-information proximity} that the proportional-sampling estimator exploits. The concept of action-information proximity will be explored more in depth in Section \ref{sec:AI_local}. 

The proof of the lower bound will be presented in Section \ref{sec:lowerBound}, consisting of the following steps.

1.  \emph{Discrete Private Learning} (Section \ref{sec:discreLearner}). We formulate a {discrete} version of the original problem where both the learner and adversary estimate the discrete index associated with a certain sub-interval that contains the target, instead of the continuous target value. The discrete framework is conceptually clearer, and will allow us to deploy information-theoretic tools with greater ease. 

2.  \emph{Localized Query Complexity} (Section \ref{sec:localComp}). Within the discrete version, we prove a localized query complexity result: conditioning on the target being in a coarse sub-interval of $[0, 1)$, any accurate learner still needs to submit a  large number of queries {within} the said sub-interval. The main argument hinges on Fano's inequality and a characterization of the conditional entropy of the queries and the target. 

3. \emph{Proportional-Sampling Estimator} (Section \ref{sec:AI_local}). We use the localized query complexity in the previous step to prove a query complexity lower bound for the discrete version of Bayesian Private Learning. This is accomplished by analyzing the performance of the family of proportional-sampling estimators, where the adversary reports index of a sub-interval that is sampled randomly with probabilities proportional to the number of learner queries each sub-interval contains. We will show that the proportional-sampling estimator will succeed with overwhelming probability whenever an accurate learner strategy submits too few queries, thus obtaining the desired lower bound. In fact, we  will prove a more general lower bound, where the learner can make mistakes with a positive probability. 

4. \emph{From Discrete to Continuous} (Section \ref{sec:DiscToCont}). We complete the proof by connecting the discrete version back to the original, continuous problem. Via a reduction argument, we show that the original query complexity is always bounded from below by its discrete counterpart with some modified learner error parameters, and the final lower bound will be obtained by optimizing over these parameters. The main difficulty in this portion of the proof is due to the fact that an accurate continuous learner estimator is insufficient for generating an accurate discrete estimator that is correct almost surely. We will resolve this problem by carefully bounding the learner's probability of estimation error, and apply the discrete query lower bound developed in the previous step, in which the learner is allowed to make mistakes.

\section{The Upper Bound}
\label{sec:upperbound}

We prove the upper bound of Theorem \ref{thm:main}.   The bound has appeared in \citet{xu2017private} and \citet{tsixuxu2017}, which proposed, without a formal proof, the Replicated Bisection  learner strategy that achieves $(\delta, L)$-privacy with   $L\log(1/\veps) - L(\log(L) -1)$ queries. For completeness, we first review the Replicated Bisection strategy and subsequently give a formal proof of its privacy and accuracy. The main idea behind Replicated Bisection is to create $L$ {identical} copies of a bisection search in a strictly symmetrical manner so that the adversary wouldn't be able to know which one of the $L$ searches is associated with the target. The strategy takes as initial inputs $\veps$ and $L$, and proceeds in two phases:
\begin{enumerate}
\item \emph{Phase 1 - Non-adaptive Partitioning}. The {learner} submits $L-1$ (non-adaptive) queries: 
 \begin{equation}
 Q_1 = \frac{1}{L},\: Q_2 = \frac{2}{L}, \ldots,\: Q_{L-1} = 1-\frac{1}{L}, 
 \end{equation}
Adjacent queries are separated by a distant of $1/L$, and together they partition the unit interval into $L$ disjoint sub-intervals of length $1/L$ each. We will refer to the interval $[(i-1)/L, i/L)$ as the \emph{$i$th sub-interval}. Because the queries in this phase are non-adaptive, after the first $L-1$ queries, while the learner knows which sub-interval contains the target, $X^*$, the adversary has gained no information about $X^*$.   We will denote by $\calI^*$ the sub-interval that contains $X$. 

\item \emph{Phase 2 - Replicated Bisection}. The second phase further consists of a sequence of $K$\emph{rounds}, $K = \log\lt(\frac{1}{L\veps}\rt)$. In each round, the learner submits one query in each of the $L$ sub-intervals, and the location of the said query relative to the left end of the sub-interval is the same across all sub-intervals. Crucially, in the $k$th round, the query corresponds to the $k$th step in a bisection search carried out in the sub-interval $\calI^*$, which contains the target. The rounds continue until the learner has identified the location of $X^*$ with sufficient accuracy within  $\calI^*$. The queries outside of $\calI^*$ serve only the purpose of obfuscation by maintaining a strict symmetry. Figure \ref{fig:pseuRBS} contains the pseudo-code for Phase 2. 

\begin{figure} 
\hrule 
\vspace{3pt}
\flushleft {\bf Replicated Bisection (Phase 2)}
\vspace{3pt}
\hrule
\vspace{3pt}
\begin{algorithmic}
\State $l^*\gets$ the index of $\calI^*$,  $K \gets \log\lt(\frac{1}{L\veps}\rt)$, $D_0 \gets \frac{1}{2L}$
\For{k:=0}{K-1} 
\Begin
\For{l:=0}{L-1}
	\State $Q_{ (k+1)L+l} \gets (l-1)\frac{1}{L}+D_k$ 
	
\vspace{3pt}

\If{$\mbox{$R_{(k+1)L+l^*}=0$ (i.e., $X^*>Q_{(k+1)L+l^*}$)  }$} 
	\State $D_{k+1} \gets D_k+\frac{1}{L}\lt(\frac{1}{2}\rt)^k$ 
\Else
	\State $D_{k+1} \gets D_k-\frac{1}{L}\lt(\frac{1}{2}\rt)^k$
\End
\vspace{5pt}
\hrule
\end{algorithmic}
\caption{Pseudo-code for Phase 2 of the Replicated Bisection strategy. $D_k$ represents the distance from a query  submitted in the $k$th round to the left end-point of its corresponding sub-interval. }
\label{fig:pseuRBS}
\end{figure}
\end{enumerate}

Denote by $Q^*$ the last query that the learner submits in the sub-interval $\calI^*$ in Phase $2$, and by $R^*$ its response. It follows by construction that either  $R^*=0$ and $X^*\in [Q^*-\veps,Q^*)$,  or $R^*=1$ and $X^*\in [Q^*,Q^*+\veps)$. Therefore, the learner can produce the estimator by setting $\what{X}$ to the mid point of either $[Q^*-\veps,Q^*)$ or $[Q^*,Q^*+\veps)$, depending on the value of $R^*$, and this guarantees of an additive error of at most $\veps/2$. We have thus shown that the Replicated Bisection strategy is $\veps$-accurate. The following result shows that it is also private; the proof is given in Appendix \ref{app:prop:upper}.

\begin{proposition}
\label{prop:upper}
Fix  $\veps$ and $\delta$ in  $ (0, 1)$ and $L\in \N$, such that $\veps < \delta/4$ and $\delta < 1/L$. The Replicated Bisection strategy is $(\delta,L)$-private. 
\end{proposition}

Finally, we verify the number of queries used by Replicated Bisection: the first phase employs $L-1$ queries, and the second phase uses $L$ queries per round, across $\log(\frac{1}{L\veps})$ rounds, leading to a total of $(L-1)+L\log(\frac{1}{L\veps})= L\log(1/\veps) - L(\log L-1)-1$ queries. This completes the proof of the query complexity upper bound in Theorem \ref{thm:main}. 

\section{The Lower Bound} 
\label{sec:lowerBound}
We prove the query complexity lower bound in Theorem \ref{thm:main} in this section. An overview of the main steps can be found in Section \ref{sec:proofOverview}. 

\subsection{Discrete Bayesian Private Learning}
\label{sec:discreLearner}

 We begin by formulating a discrete version of the original problem, where the goal for both the learner and the adversary is to recover a discrete index associated with the target, as opposed to generating a continuous estimator. We first create two nested partitions of the unit interval consisting of equal-length sub-intervals, where one partition is coarser than the other. The objective of the learner is to recover the index associated with the sub-interval containing $X^*$ in the finer partition, whereas that of the adversary is to recover the target's index corresponding to the coarser partition (an easier task!). We consider this discrete formulation because it allows for a simpler analysis using Fano's inequality,  setting the stage for the localized query complexity lower bound in the next section. 

\begin{figure}[h]
\centering
\includegraphics[scale=.6]{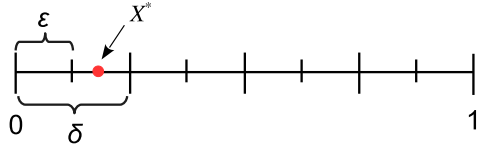}
\caption{An example of the indices $J(\cdot, X^*)$. Here, $\delta=0.2$ and $\veps=0.1$, and the target $X^*=0.15$. The target thus belongs to the first sub-interval in a $\delta$-uniform partition, and the second sub-interval in an $\veps$-uniform partition. We have that $J(\delta, X^*)=1$ and $J(\veps, X^*)=2$.}
\label{fig:ills1}
\end{figure}

 Formally, fix $s \in (0,1)$ such that $1/s$ is an integer. Define $M_i(s)$ to be the sub-interval
\begin{equation}
M_s(i) = [(i-1)s, is), \quad i = 1, 2, \ldots, 1/s. 
\end{equation} 
In particular, the set $\calM_s := \{M_s(i): i=1, \ldots, 1/s \}$ is a partition of $[0, 1)$ into $1/s$  sub-intervals of length $s$ each. We will refer to $\calM_s$ as the $s$-uniform partition.  Define
\begin{align}
J(s, x) = j, \quad \mbox{ s.t. } x \in M_s(j). 
\end{align}
That is, $J(s,x)$ denotes the indices of the interval containing $x$ in the $s$-uniform partition. A visualization of the index $J(\cdot, \cdot)$ is given in Figure \ref{fig:ills1}. 

We now formulate an analogous, and slightly more general, definition of accuracy and privacy for the discrete problem.  We will use the super-script $D$ to distinguish them from their counterparts in the original, continuous formulation. Just like the learner strategy in Definition \ref{def:learnerContinuous}, a discrete learner strategy, $\phi^D$, is allowed to submit queries at any point along $[0, 1)$, and has access to the random seed, $Y$. The only difference is that, instead of generating a continuous estimator, a discrete learner strategy produces an estimator for the \emph{index} of the sub-interval containing the target in an $\veps$-uniform partition, $J(\veps,X^*)$. 
\begin{definition}[$(\veps,\nu)$-accuracy - Discrete Version] Fix $\veps$ and $\nu \in (0,1)$. A discrete learner strategy, $\phi^D$, is $(\veps, \nu)$-accurate if it produces an estimator, $\what{J}$, such that 
\begin{equation}
\pb\lt(\what{J} \neq J(\veps, X^*)  \rt) \leq \nu. 
\end{equation}
\end{definition}
Importantly, in contrast to its continuous counterpart in Definition \ref{def:eps-acc-cont} where the estimator must satisfy the error criterion with probability one, the discrete learner strategy is allowed to make mistakes up to a probability of $\nu$. 

The role of adversary is similarly defined in the discrete formulation: upon observing all $n$ queries, the adversary generates an estimator, $\what{J}^a$, for the index associated with the sub-interval containing $X^*$ in the (coarser) $\delta$-uniform partition, $J(\delta,X^*)$. The notion of $(\delta,L)$-privacy for a discrete learner strategy is defined in terms of the adversary's (in)ability to estimate the index $J(\delta, X^*)$.

\begin{definition}[$(\delta, L)$-privacy - Discrete Version] Fix $\delta \in (0,1)$ and $L \in \N$. A discrete learner strategy, $\phi^D$, is $(\delta,L)$-private if under any adversary estimator $\what{J}^a$, we have that
\begin{equation}
\pb\lt( \what{J}^a = J(\delta, X^*) \rt) \leq 1/L. 
\end{equation}
We will denote by $\Phi^D_n$ as the family of discrete learner strategies that employ at most $n$ queries. 
\end{definition}
We are now ready to define the query complexity of the discrete formulation, as follows:   
 \begin{equation}
 N^D(\veps,\nu, \delta, L) = \min\{n: \, \Phi^D_n \mbox{ contains a   strategy that is both $(\veps, \nu)$-accurate and $(\delta, L)$-private}\}.\nonumber
 \end{equation}
A main result of this subsection is the following lower bound on $N^D$, which we will convert into one for the original problem in Section \ref{sec:DiscToCont}. 

\begin{proposition}[Query Complexity Lower Bound for Discrete Learner Strategies]
\label{prop:disc_lowerbound}
Fix  $\veps$, $\nu$ and $\delta$ in  $ (0, 1)$ and $L \in \N$, such that $\veps < \delta < 1/L$. We have that
\begin{equation}
N^D(\veps, \nu, \delta, L) \geq L\lt[ (1-\nu) \log(\delta/\veps) - h(\nu)\rt], 
\end{equation}
where $h(p)$ is the Shannon entropy of a Bernoulli random variable with mean $p$: $h(p) = -p \log(p) - (1-p)\log(1-p)$ for $p\in (0,1)$,  and $h(0)=h(1)=0$. 
\end{proposition}

\subsection{Localized Query Complexity Lower Bound} 
\label{sec:localComp}

We prove Proposition \ref{prop:disc_lowerbound} in the next two subsections. The first step, accomplished in the present subsection, is to use Fano's inequality to establish a query complexity lower bound localized to a sub-interval: conditional on the target belonging to a sub-interval in the $\delta$-partition,  any discrete-learner strategy must devote a non-trivial number of queries  in that sub-interval if it wishes to be reasonably accurate. Since all learner strategies considered in the next two subsections will be for the discrete problem,  we will refer to them simply as learner strategies when there is no ambiguity.

Fix $n\in \N$, and a learner strategy $\phi^D \in \Phi^D_n$. Because the strategy will submit at most $n$ queries, without loss of generality, we may assume that if the learner wishes to terminate the process after the first $K$ queries, then she will simply set $Q_i$ to $0$ for all  $i \in \{K+1, K+2, \ldots, n\}$, and the responses for those queries will be trivially equal to $0$ almost surely. Denote by $\calQ^j$ the set of queries that lie within the sub-interval $M_\delta(j)$: 
\begin{equation}
\calQ^j \bydef \{Q_i, \ldots, Q_n\} \cap   M_\delta(j),
\label{eq:Qjdef} 
\end{equation}
and by $|\calQ^j|$ its cardinality. Denote by $\calR^j$ the set of responses for those queries in $\calQ^j$. Define $\xi_{j,y}$ to be the learner's  (conditional) probability of error: 
\begin{equation}
\xi_{j,y} = \pb\lt(\what{J} \neq J(\veps, X^*) \bbar J(\delta, X^*) = j , Y = y\rt), \quad  j \in \{1,\ldots, 1/\delta\}, \, y\in [0, 1). 
\end{equation}
We have the following lemma. The proof is based on Fano's inequality. 

\begin{lemma}[Localized Query Complexity]
\label{lem:EQlower} 
 Fix $\veps$, $\nu$ and $\delta \in (0,1)$, $\veps<\delta$, and an $(\veps, \nu)$-accurate discrete learner strategy. We have that
\begin{equation}
\E\lt(|\calQ^j| \bbar J(\delta, X^*) = j, Y=y\rt) \geq (1-\xi_{j,y})\log(\delta/\veps)- h(\xi_{j,y}), 
 \label{eq:EQj}
\end{equation}
for all $j \in \{1,\ldots, 1/\delta\}$, and $y\in [0, 1)$, and 
\begin{equation}
  \sum_{j=1}^{{1}/{\delta}} \delta \int_{0}^1 \xi_{i,y} \, dy \leq \nu. 
 \label{eq:xitoAvg}
\end{equation}
\end{lemma}

\bpf Denote by $\calE_{j,y}$ the event:
\begin{equation}
\calE_{j,y} = \{J(\delta, X^*) = j, Y=y\}. 
\end{equation} 
Because the random seed, $Y$, is uniformly distributed over $[0, 1)$, Eq.~\eqref{eq:xitoAvg} follows directly from the learner strategy's being $(\veps,\nu)$-accurate: 
\begin{equation}
\nu\geq \pb\lt(\what{J}\neq J(\veps, X^*) \rt) = \sum_{j=1}^{1/\delta} \int_{0}^1 \pb(\calE_{j,y})\pb\lt(\what{J}\neq J(\veps, X^*) \bbar \calE_{j,y} \rt) \, dy =  \sum_{j=1}^{1/\delta}  \delta \int_{0}^1  \xi_{j,y} \, dy.
\end{equation}

We now show Eq.~\eqref{eq:EQj}. Fix $j \in \{1,\ldots, 1/\delta\}$, and $y\in [0, 1)$.  
We begin by making the simple observation that, conditional on $\calE_{j,y}$, the subset of queries $\calQ^j$ together with their responses $\calR^j$ is sufficient for generating the learner's estimator, $\what{J}$, because under this conditioning, any query that lies outside the sub-interval $M_\delta(j)$ provides no additional information about the location of $X^*$ than what is already known. Furthermore, since the random seed $Y$ is fixed to $y$, the  $i$th query, $Q_i$, is a deterministic function of the first $i-1$ responses. We conclude that the set of responses $\calR^j$ alone is sufficient for generating $\what{J}$.

For an event, $\calE$, we will denote by $H(A|B,\calE)$ the conditional entropy $H(A \bbar B)$ under the probability law $\pb(\cdot | \calE)$:
\begin{equation}
H(A|B,\calE) \bydef - \sum_{a \in \calA, b \in \calB} \pb(A=a, B=b \bbar \calE) \log \lt(\pb(A=a\bbar B=b,\calE) \rt), 
\label{eq:condEDef1}
\end{equation}
where $\calA$ and $\calB$ are the alphabets for random variables $A$ and $B$, respectively. Similarly, define
\begin{equation}
H(A\bbar \calE)  \bydef - \sum_{a \in \calA} \pb(A=a \bbar \calE) \log \lt(\pb(A=a\bbar\calE) \rt). 
\label{eq:condEDef2}
 \end{equation} 
Let $V \in \{0,1\}^n$ be the vector representation of $\calR^j$:
\begin{equation}
V_i  =  \mbox{the $i$th element of $\calR^j$}, \quad i =1, 2, \ldots, |\calQ^j|, 
\end{equation}
and $V_i  = 1$ for all $i = |\calQ^j|, |\calQ^j|+1, \ldots, n$.  The conditional entropy of $V$  given $\calE_{j,y}$ satisfies: 
\begin{align}
H\lt( V \bbar \calE_{j,y} \rt) =& \sum_{k=1}^n  H\lt( V \bbar \calE_{j,y},|\calQ^j| = k \rt)\pb\lt(|\calQ^j| = k \bbar \calE_{j,y}\rt) \nln
\leq  & \sum_{k=1}^n  k\pb\lt(|\calQ^j| = k \bbar \calE_{j,y}\rt) \nln
= & \E\lt(|\calQ^j| \bbar \calE_{j,y} \rt),
\label{eq:entropV}
\end{align}
where the inequality follows from the fact that, conditional on there being $k$ responses in $\calR^j$, we know that only the first $k$ bits of $V$ can be random, and hence the entropy of $V$ cannot exceed $k$, which is the entropy of a length-$k$ vector where each entry is an independent Bernoulli random variable with mean $1/2$.  We now invoke the following lemma by Robert Fano (cf.~Section 2.1 of \citet{cover2012elements}).
\begin{lemma}[Fano's Inequality] Let $A$ and $B$ be two random variables, where $A$ takes values in a finite set, $\calA$. Let $\what{A}$ be a discrete random variable taking values in $\calA$, such that $\what{A} = f(B,C)$, where $f$ is a deterministic function, and $C$ a random variable independent from both $A$ and $B$. Let  $p = \pb(\what{A} \neq A)$. We have that
\begin{equation}
H\lt(A \bbar B \rt) \leq h(p) + p\lt(\log|\calA|-1\rt), 
\label{eq:fanos}
\end{equation}
where $H(A \bbar B)$ is the conditional entropy of $A$ given $B$. 
\end{lemma}

We apply Fano's inequality with the substitutions: $A \leftarrow J(\veps, X^*)$, $B \leftarrow V$,  and $\what{A} \leftarrow \what{J}$. Eq.~\eqref{eq:fanos} yields 
\begin{equation}
H\lt(J(\veps, X^*) \bbar V, \calE_{j,y}\rt)  \leq h(\xi_{j,y}) + \xi_{j,y}\log(\delta/\veps), 
\label{eq:Hless0}
\end{equation}
where we have used the fact that, conditional on the event $\calE_{j,y}$, the index $J(\veps, X^*)$ can take at most $\delta/\veps$ values. By the chain rule of conditional entropy, we have that
\begin{align}
H\lt(J(\veps, X^*) \bbar V, \calE_{j,y}\rt) =& H\lt(J(\veps, X^*), V \bbar \calE_{j,y}\rt)- H\lt(V\bbar \calE_{j,y}\rt) \nln
\geq & H\lt(J(\veps, X^*) \bbar \calE_{j,y}\rt) - H\lt(V\bbar \calE_{j,y}\rt) \nln
\sk{a}{=} & \log(\delta/\veps)-  H\lt(V\bbar \calE_{j,y}\rt)  \nln
\sk{b}{\geq}& \log(\delta/\veps)-   \E\lt(|\calQ^j| \bbar \calE_{j,y} \rt), 
\label{eq:condiEntropJ} 
\end{align}
where step $(a)$ follows from the fact that conditional on $\calE_{j,y}$, $J(\veps, X^*)$ is  uniformly distributed over $\delta/\veps$ possible values, and step $(b)$ follows from Eq.~\eqref{eq:entropV}. Combining Eqs.~\eqref{eq:Hless0} and \eqref{eq:condiEntropJ} yields
\begin{align}
 \E\lt(|\calQ^j| \bbar \calE_{j,y} \rt) \geq & \log(\delta/\veps) - H\lt(J(\veps, X^*) \bbar V, \calE_{j,y}\rt)  \nln
\geq & \log(\delta/\veps) - \lt( h(\xi_{j,y}) + \xi_{j,y} \log(\delta/\veps) \rt) \nln
=   & (1-\xi_{j,y})\log(\delta/\veps) - h(\xi_{j,y}). 
\end{align}
This proves Lemma \ref{lem:EQlower}. \qed. 

\subsection{Proportional-Sampling Estimator}
\label{sec:proportionalSample}
We now use the local complexity result in Lemma \ref{lem:EQlower} to complete the proof of Proposition \ref{prop:disc_lowerbound}. The lemma states  that if the target were to lie in a given sub-interval in the $\delta$-uniform partition, then an accurate learner strategy would have to place at least $\log(\delta/\veps)$ queries within the said sub-interval on average. A naive approach to using the lemma would let the adversary's estimator select a sub-interval among those  in $\calM_\delta$ that contain approximately $\log(\delta/\veps)$ queries, with the hope that the learner would have to induce $L$ such sub-intervals to ensure privacy. However, for this estimator to be a credible threat,  the local complexity lower bound needs to hold with almost certainty, instead of merely on average. Unfortunately, it may not be possible to strengthen the lower bound to holding with certainty in general. For instance, it is easy to see that the learner could ``guess'' the target's location with only two queries, at the expense of possibly increasing the average number of queries overall. This suggests that forcing the number of queries to concentrate at, or above, $\log(\delta/\veps)$ under an arbitrary learner strategy is likely very difficult, if not impossible.  

Nevertheless, we are still able to take advantage of the average local complexity in Lemma \ref{lem:EQlower}, but with a more robust adversary estimator. We will focus on a family of \emph{proportional-sampling} estimators for the adversary. Despite their simplicity, these estimators prove to be sufficiently powerful against any learner strategy that uses a ``small'' number of queries. 
\begin{definition} 
\label{def:psample}
A {proportional-sampling estimator}, $\what{J}^a$, is generated according to the distribution: 
\begin{equation}
\pb\lt(\what{J}^a = j\rt) = \frac{|\calQ^j|}{\sum_{j'=1}^{1/\delta} |\calQ^{j'}|}, \quad j = 1, 2, \ldots, 1/\delta. 
\end{equation}
That is, an index is sampled with a probability proportional to the number of queries that fall within the corresponding sub-interval in the $\delta$-uniform partition. 
\end{definition}

We next bound the probability of correct estimation when the adversary employs a proportional-sampling estimator: for all $j= 1, 2, \ldots, 1/\delta$, we have that
\begin{align}
\pb\lt( \what{J}^a =  J(\delta, X^*) \bbar  \calE_{j,y}\rt)  = & \pb\lt( \what{J}^a = j \bbar  \calE_{j,y} \rt) \nln
= & \E\lt( \lt. \frac{|\calQ^j|}{\sum_{j'=1}^{1/\delta} |\calQ^{j'}|} \, \,  \rt| \,\,  \calE_{j,y} \rt) \nln
\sk{a}{=} & \frac{1}{n} \E\lt( |\calQ^j| \bbar  \calE_{j,y} \rt)  \nln
\sk{b}{\geq} & \frac{1}{n}\lt( (1-\xi_{j,y}) \log(\delta/\veps) - h(\xi_{j,y})\rt), 
\label{eq:jatoq}
\end{align}
where step $(a)$ follows from the fact that $\phi^D \in \Phi^D_n$, and hence $\sum_{j'=1}^{1/\delta} |\calQ^{j'}| = n$, and step $(b)$ from Lemma \ref{lem:EQlower}.  Recall that the learner strategy is $(\veps, \nu)$-private, and the random seed $Y$ has a probability density of $1$ in $[0, 1)$ and zero everywhere else.
Since Eq.~\eqref{eq:jatoq} holds for all $j$ and $y$, we can integrate and obtain the adversary's overall probability of correct estimation:
\begin{align}
\pb\lt( \what{J}^a =  J(\delta, X^*) \rt) =&  \sum_{j=1}^{1/\delta} \int_0^1 \pb(\calE_{j,y})\pb\lt(\what{J}^a =  J(\delta, X^*) \bbar \calE_{j,y} \rt) \, dy \nln
\sk{a} \geq & \frac{1}{n} \sum_{j=1}^{1/\delta}  \int_0^1 \delta \lt[ (1-\xi_{j,y}) \log(\delta/\veps) - h(\xi_{j,y})\rt] \, dy \nln
= & \frac{1}{n}\lt[ \lt(1-\sum_{j=1}^{1/\delta} \delta  \int_0^1  \xi_{j,y}  \, dy \rt) \log(\delta/\veps) -    \sum_{j=1}^{1/\delta}   \delta  \int_0^1 h(\xi_{j,y})  \, dy \rt] \nln
\sk{b} \geq & \frac{1}{n}\lt[ (1-\nu) \log(\delta/\veps) -    \sum_{j=1}^{1/\delta}   \delta  \int_0^1 h(\xi_{j,y})  \, dy \rt] \nln
\sk{c}{\geq} & \frac{1}{n}\lt[ (1-\nu) \log(\delta/\veps) - h\lt(\sum_{j=1}^{1/\delta} \delta  \int_0^1  \xi_{j,y}  \, dy\rt)\rt] \nln
\sk{d}{\geq}& \frac{1}{n}\lt[ (1-\nu) \log(\delta/\veps) - h(\nu)\rt], 
\label{eq:propEstPlower}
\end{align}
where step $(a)$ follows from Eq.~\eqref{eq:jatoq},  steps $(b)$ and $(d)$ from Eq.~\eqref{eq:xitoAvg} of Lemma \ref{lem:EQlower}, i.e.,
\begin{equation}
  \sum_{j=1}^{{1}/{\delta}} \delta \int_{0}^1 \xi_{i,y} \, dy \leq \nu. 
\end{equation}
Step $(c)$ is a result of Jensen's inequality and the Bernoulli entropy function $h(\cdot)$'s being concave. 

Recall that, in order for a learner strategy to be $(\delta,L)$-private, we must have that $\pb\lt( \what{J}^a =  J(\delta, X^*) \rt)  \leq \frac{1}{L}$ for \emph{any} adversary estimator $\what{J}^a$. Eq.~\eqref{eq:propEstPlower} thus implies that 
\begin{equation}
n \geq L\lt[ (1-\nu) \log(\delta/\veps) - h(\nu)\rt], 
\end{equation}
 is necessary. Because this holds for any accurate and private learner policy, we have thus proven Proposition \ref{prop:disc_lowerbound}. 

\subsection{From Discrete to Continuous Strategies}
\label{sec:DiscToCont}
We now connect Proposition \ref{prop:disc_lowerbound} to the original continuous estimation problem. The next proposition is the main result of this subsection. The core of the proof is a reduction that constructs a $(\beta\veps,\beta^{-1})$-accurate and $(\delta,L)$-private discrete learner strategy from an $\veps$-accurate and $(\delta,L)$-private continuous learner strategy. 
\begin{proposition} 
\label{prop:dis_to_cont}
Fix  $\veps$ and $\delta$ in  $ (0, 1)$ and $L \in \N$, such that $\veps < \delta/4$ and $\delta < 1/L$. Fix $\beta \in [2 \, , \,    {\delta}/{\veps}]$.\footnote{To avoid the use of rounding in our notation, we will assume that $\delta$ is an integer multiple of $\beta \veps$. } We have that
\begin{equation}
 N (\veps, \delta, L) \geq N^D (\beta\veps, \beta^{-1}, \delta, L). 
\end{equation}
\end{proposition}

\bpf Fix $n\in \N$ and a continuous learner strategy, $\phi \in \Phi_n$,   such that $\phi$ is both $\veps$-accurate and $(\delta, L)$-private. Let $\what{X}$ the estimator of $\phi$. It suffices to show that there exists a function $f: [0,1) \to \calM_{\beta\veps}$, such that by using the same queries as $\phi$, and setting $\what{J} = f(\what{X})$ we obtain a $(\beta\veps, \beta^{-1})$-accurate and $(\delta, L)$-private discrete learner strategy. Specifically, let $\phi^D$ be the discrete learner strategy that submits the same queries as $\phi$, and produces the estimator 
\begin{equation}
\what{J} = J(\beta\veps,\what{X}). 
\end{equation}
That is, $\what{J}$ reports the index of the sub-interval in the $\beta\veps$-uniform partition that contains the continuous estimator, $\what{X}$.  

We first show that the induced discrete learner strategy is $(\beta\veps, \beta^{-1})$-private. The intuition is that if the target $X^*$ is sufficiently far away from the edges of the sub-interval in the $(\beta\veps)$-uniform partition to which it belongs, then both $X^*$ and $\what{X}$ will belong to the same sub-interval, and we will have $J(\beta\veps,\what{X}) = J(\beta\veps,X^*)$. To make this precise, denote by $\calG_{\beta\veps}$ the set of end points of the sub-intervals in the $(\beta\veps)$-uniform partition: $\calG_{\beta\veps} \bydef \{0, \beta\veps, 2\beta\veps, \ldots, 1-\beta\veps, 1\}.$ Let $\calS$ be the set of all points in $[0,1)$ whose distance to $\calG_{\beta\veps}$ is greater than $\veps/2$: 
\begin{equation}
\calS = \{x \in [0, 1): \min_{y \in \calG_{\beta\veps}}|x-y|> \veps/2\}. 
\end{equation}
It is not difficult to show that the  Lebesgue measure of $\calS$ satisfies $\mu^\calL(\calS) = {\veps}/{(\beta\veps)} = \beta^{-1},$ where $\veps$ is the length of the intersection of $\calS$ with each of the $(\beta\veps)^{-1}$ sub-intervals in a $(\beta\veps)$-partition. 
Since $\phi$ is $\veps$-accurate, we know that $\what{X}$ must be no more than $\veps/2$ away from $X^*$, and hence $\what{J} = J\lt(\beta \veps, X^* \rt)$ whenever $X^*\notin \calS$, which implies
\begin{align}
\pb\lt(\what{J} \neq J\lt(\beta \veps, X^* \rt) \rt) \leq \pb\lt( X^* \in \calS \rt)  =\mu^\calL(\calS) =  \beta^{-1}. 
\label{eq:probContToDis}
\end{align}
This shows that $\phi^D$ is $(\beta\veps, \beta^{-1})$-accurate.  

We next show that $\phi^D$ is also $(\delta, L)$-private. For the sake of contradiction, suppose otherwise. Then, there exists an estimator for the adversary, $\what{J}^a$, such that
\begin{equation}
\pb\lt(\what{J}^a = J(\delta, X^*) \rt) > 1/L. 
\label{eq:estimatContra}
\end{equation}
We now use $\what{J}^a$ to construct a ``good'' adversary estimator for the continuous version: let $\what{X}^a$ be the mid point of the sub-interval $M_\delta(\what{J}^a)$, where $M_\delta(j)$ is the $j$th sub-interval in the $\delta$-uniform partition.  If $\what{J}^a = J(\delta, X^*)$, then $M_\delta(j)$ contains $X^*$, and since the length of $M_\delta(j)$ is $\delta$, we must have $\lt|\what{X}^a - X^*\rt| \leq \delta/2$, and from Eq.~\eqref{eq:estimatContra}, this implies
\begin{equation}
\pb\lt(\lt| \what{X}^a - X^* \rt| \leq \delta/2 \rt) > 1/L. 
\end{equation}
We therefore conclude that if  an estimator satisfying Eq.~\eqref{eq:estimatContra} did exist, then the original continuous learner strategy, $\phi$, could not have been $(\delta,\veps)$-private, which leads to a contradiction. We have thus shown  that $\phi^D$ is $(\beta\veps, \beta^{-1})$-accurate and  $(\delta, L)$-private. Because $\phi^D$ uses the same sequence of queries as $\phi$, we conclude that $ N (\veps, \delta, L) \geq N^D (\beta\veps, \beta^{-1}, \delta, L)$.  This proves Proposition \ref{prop:dis_to_cont}.  \qed

\subsubsection{Completing the Proof of the Lower Bound} We are now ready to establish the query complexity lower bound in Theorem \ref{thm:main}. Fix  $\veps$ and $\delta$ in  $ (0, 1)$ and $L \in \N$, such that $\veps < \delta/4$ and $\delta < 1/L$.   Using  Propositions \ref{prop:disc_lowerbound} and \ref{prop:dis_to_cont}, we have that for any $\beta \in [2, \delta/\veps]$, 
\begin{align}
N(\veps, \delta, L) \geq & N^D (\beta\veps, \beta^{-1}, \delta, L) \sk{b}{\geq}   L\lt[ (1-\beta^{-1}) \log\lt(\frac{\delta}{\veps} \beta^{-1}\rt) - h(\beta^{-1})\rt],
\end{align}
where the last step follows from Proposition \ref{prop:disc_lowerbound} by substituting $\nu$ with $\beta^{-1}$ and $\veps $ with $\beta \veps$. Letting $\gamma \bydef \beta^{-1}$, 
the above inequality can be rearranged to become
\begin{align}
\frac{N(\veps, \delta, L) }{L}\geq &   (1-\gamma) \log\lt(\frac{\delta}{\veps} \gamma\rt) - h(\gamma)\nln
\sk{a}{\geq} &   (1-\gamma) \log\lt(\frac{\delta}{\veps} \gamma\rt) + 2(1-\gamma)\log(\gamma) \nln
= & \log(\delta/\veps) - \gamma \log(\delta/\veps) +  3(1-\gamma)\log \gamma \nln
\geq  & \log(\delta/\veps) - \gamma \log(\delta/\veps) +  3\log \gamma, 
\label{eq:NoverL}
\end{align}
where step $(a)$ follows from the assumption that $\gamma = \beta^{-1} \leq 1/2$, and the fact that $h(x) \leq -2(1-x)\log(x)$ for all $x\in (0, 1/2]$. Consider the choice: 
\begin{equation}
\beta = \log(\delta/\veps). 
\end{equation}
To verify $\beta$ still belongs to the range $[2,\delta/\veps]$, note that the assumption that $\veps < \delta/4$ ensures  $\beta \geq 2$, and because $x>\log(x)$ for all $x>0$, we have that $\beta<\delta/\veps$.  Substituting $\gamma$ with $(\log(\delta/\veps))^{-1}$ in Eq.~\eqref{eq:NoverL}, we have that
\begin{align}
\frac{N(\veps, \delta, L) }{L} \geq & \log(\delta/\veps) - 1 - 3\log\log(\delta/\veps)
\end{align}
or, equivalently,  $N(\veps, \delta, L) \geq L\log(1/\veps) - L\log(2/\delta)- 3L\log\log(\delta/\veps). $ This completes the proof of the lower bound in Theorem \ref{thm:main}. 

\section{Generalizations}
\label{sec:generalizations}

The aim of the present section is to explore the generality of the our proof technique by extending the main result, Theorem \ref{thm:main}, to two variants of the original model. Before delving into the technical development, we will take a closer look at the workhorse behind the lower bound proof, namely, the proportional-sampling estimators, in order to better understand {what features of an active learning problem make  its privatization especially costly in terms of query complexity.
} 

\subsection{Action-Information Proximity}
\label{sec:AI_local}

As was alluded to in Section \ref{sec:proofOverview}, the proportional-sampling estimator's efficacy stems from the fact that,  in order to accurately locate the target, a {large} number of the queries must be {spatially close} to the said target. This means that the learner is essentially forced to devote a substantial number of queries in multiple, separate locations, such as is the case with the Replicated Bisection strategy in Section \ref{sec:upperbound}. Otherwise, a disproportionally large fraction of the queries will concentrate around the target, thus exposing the learner to the sort of attacks from a proportional-sampling estimator that exploit the queries' non-uniform spatial arrangement. 

  Extending the above ideas further, we see that the culprit in the learner's plight is a strong co-location of {actions} (e.g., queries) and {information} (e.g., location of target), to be dubbed \emph{action-information proximity}: queries only reveal information when the target is situated nearby, and far-away queries provide little, if any, hint about the target's precise location.  We may postulate that, in general, privacy will come at a steep cost in those learning problems where: 
\begin{enumerate}
\item  Learning requires a large number of actions, regardless of the level of privacy. (This requirement corresponds to the local complexity result of Lemma \ref{lem:EQlower} in our problem.)
\item According to some metric of distance, far-away actions are essentially uninformative. 
\end{enumerate}
If both conditions hold, then whenever the learner uses relatively few learning actions, these actions are likely forced to correlate with, and thus compromise, the very information that the learner would like to conceal. 

In the remainder of this section, we will explore the applicability of the concept of action-information proximity by studying two generalizations of the original model. In the first, we consider a high-dimensional version of Bayesian Private Learning, where the target lies in a $d$-dimensional cube and the queries correspond to affine hyperplanes. Unlike in one dimension, a single query can now solicit information from {multiple} regions within the unit cube, and thus weakening the action-information proximity. In the second variant, we ask what happens when the adversary only has partial observations of the queries. The partial observation assumption indirectly weakens the link between action and information, because a query that would have been revealing about the target may not be fully observed by the adversary. We will see that the proof techniques developed in the proceeding sections can still provide non-trivial characterizations of query complexity. Unfortunately, these characterizations will be weaker compared to those in Theorem \ref{thm:main}, possibly a result of the weakened action-information proximity.

\subsection{High-Dimensional Linear Queries}

We consider in this subsection a high-dimensional generalization of the one-dimensional Bayesian Private Learning in Section \ref{sec:intro}. Fix $d\in \N$. Let the target, $X^* = (X^*(1), \ldots, X^*(d))$, be a point distributed uniformly at random in the $d$-dimensional cube $[0, 1)^d$. A learner in search of $X^*$ is allowed to submit vector-valued queries, where 
\begin{equation}
Q_i = (Q_i(1), \ldots, Q_i(d+1)) \subset \R^d,   
\end{equation}
for which she will receive the response
\begin{equation}
R_i= \mathbb{I}\lt(\lt< Q_i(1:d), X^*\rt>\leq Q_i(d+1) \rt),
\end{equation}
where $Q_i(1:d)\bydef (Q_i(1), \ldots, Q_i(d))$ and $\lt< Q_i(1:d), X^*\rt>$ denotes the inner product $\sum_{k=1}^d Q_i(k)X^*(k)$. In simple terms, the binary response corresponds whether or not the target lies on the one side of the affine hyperplane associated with $Q_i$. Analogous to the original formulation, a learner strategy is $\veps$-accurate if it is capable of producing an estimator $\what{X}$ such that
\begin{equation}
\pb(\|\what{X} - X^*\|_{\infty}\leq \veps/2 ) = 1, 
\end{equation}
where $\|\cdot\|_\infty$ denotes the $L_\infty$ norm: $\|x\|_\infty = \max_{1\leq k \leq d}|x_k|$. The notion of $(\delta,L)$-privacy is similarly defined, by replacing the absolute value in Definition \ref{def:dLprivacy} with the $\|\cdot\|_\infty$ norm. Note that, this formulation is more general than the one-dimensional model, which can be recovered by setting  $d=1$ and $Q_i(1)=1$. 

 We have the following generalization of Theorem \ref{thm:main}. 
\begin{theorem}[Query Complexity of Private Learning with Hyperplanes] 
\label{thm:main-ddim}
Fix $d\in \N$, $d\geq 2$. There exists a constant $\delta_1>0$, such that whenever $L\geq 2^d \wedge {1}/{\delta_1}$,  $\delta < L^{-1/d}$ and  $\veps \in (0, \delta/(2d))$, the following is true:\footnote{Note that here we require $L\geq 2^d$. This is due to a natural extension of the Replicated Bisection strategy where we partition the $d$-dimensional unit cube into $L$ equal-size cubes. In order for such a partition to be well defined, $L^{1/d}$ needs to be an integer greater than or equal to $2$. It is possible to remove such a requirement on $L$ in the upper bound by using an agent strategy that only privatize along one dimension of the problem. However, that scheme will incur a higher query complexity for large $L$ and also demand a stricter requirement for $\delta$ to be at most $1/L$, instead of $L^{-1/d}$.}
\begin{enumerate}
\item Upper bound: 
\begin{equation}
 N(\veps, \delta, L) \leq   dL\log(1/\veps) - L\log L + dL^{1/d} \leq dL\log(1/\veps) - L(\log L -1). 
\end{equation}
\item Lower bound:\footnote{ Using an argument similar to that in Lemma \ref{lem:EQlower}, it can be shown that $N(\veps, \delta, L)$ must also be on the order of at least $d\log(1/\veps)$, which corresponds to the (non-private) query complexity when $L=1$. Unfortunately, retaining the joint multiplicative dependence on both $d$ and $L$ in the lower bound proves to be challenging. }
\begin{equation}
N(\veps, \delta, L)\geq c_1 \delta^{d-1} L \log(\delta/\veps) - c_2 L, 
\end{equation}
where $c_1, c_2$ are constants that can depend on $d$ but not other parameters. 
\end{enumerate}
\end{theorem}
Zooming in on the regime where $\veps$ is small compared to other parameters, we obtain the following corollary of Theorem \ref{thm:main-ddim}. 
\begin{corollary} 
\label{cor:main-ddim}
Fix $d\in \N$, $d\geq 2$. Suppose we have that  $L\geq 2^d \wedge {1}/{\delta_1}$ and  $\delta < 1/L^{1/d}$, where $\delta_1$ is defined in Theorem \ref{thm:main-ddim}. Then, there exist positive constants $c'$ and $\veps_1$, such that
\begin{equation}
c' L \log(1/\veps) \leq N(\veps, \delta, L)  \leq  dL\log(1/\veps), \quad \mbox{for all $\veps \in (0, \veps_1)$}. 
\end{equation}
The constants $c'$ and $\veps_1$ can depend on $d$ and $\delta$ but not other parameters. 
\end{corollary}

Comparing Theorem \ref{thm:main-ddim} and Corollary \ref{cor:main-ddim} to their one-dimensional counterparts in  Theorem \ref{thm:main} and Corollary \ref{cor:main}, we notice that both the upper and lower bounds have become weaker as a result of an increase in dimensionality. Nevertheless, we have been able to retain the {same} multiplicative dependence on the level of privacy, $L$, suggesting that it is a fundamental property of the problem, independent of the dimensionality. The new lower bound is non-trivial since, as we discuss in the proof that follows, the increase in dimensionality makes the task of the adversary more difficult, and as such, the one-dimensional lower bound derived in Theorem \ref{thm:main} does not apply. 

We now prove Theorem \ref{thm:main-ddim}. Throughout this section, when appropriate, we will refer to earlier proof arguments to avoid repetition and instead focus on highlighting the modifications.  We begin with the query complexity upper bound. Since an accurate learner strategy must be able to achieve a small error along each of the $d$ coordinates of $X^*$, a natural generalization of the bisection search strategy would simply conduct a separate search in each dimension, $k=1, \ldots, d$, by setting $Q_i(d+1)$ equal to the location of the query that corresponds to the bisection search along the $k$th coordinate, $Q_i(k)=1$, and $Q_i(j)=0$ for all $j\neq k$ or $d+1$. With $\log(1/\veps)$ queries per coordinate, this leads to a total of $d\log(1/\veps)$ queries. We can generalize the Replicated Bisection strategy of Section \ref{sec:upperbound} in a similar manner. For $s\in (0,1)$ such that $s^{-1}\in \N$, define the $d$-dimensional $s$-uniform partition, $\calM_{s}= \{M_s(l_1, \ldots, l_d)  \}_{(l_1, \ldots, l_d) \in \{1, \ldots, s^{-1}\}^d}$, where $M_s(l_1, \ldots, l_d)$ is the cube: 
\begin{equation}
M_s(l_1, \ldots, l_d)  = [(l_1-1)s \, ,\,  l_1s)\times \cdots \times [(l_d-1)s, l_d s).
\label{eq:ddimuniformpart}
\end{equation}
We will refer to $M_s(l_1, \ldots, l_d)$ as a {sub-cube} of the partition.  To implement the Replicated Bisection strategy with privacy level $L$, we will need to create $L$ sub-cubes, and this corresponds to a ${L^{1/d}}$-uniform partition of the unit cube, $\calM_{L^{-1/d}}$, where each sub-cube has an edge length of $L^{-\frac{1}{d}}$. 
In Phase 1, the learner submits  $dL^{1/d}$ deterministic queries to find out which sub-cube within $\calM_{L^{-1/d}}$ contains the target, where the queries take the form
\begin{equation}
(Q_i(1), \ldots, Q_i(d))= e_k  , \, Q_i(d+1) = L^{-1/d}j , \quad j=0,1, \ldots  L^{1/d}-1, \, k = 1, \ldots, d, 
\end{equation}
where $e_k$ is the $d \times 1$ vector with $1$ in the $k$th coordinate, and $0$ everywhere else. In Phase 2, the learner conducts a copy of a bisection search  in each of the $L$ sub-cubes of $\calM_{L^{-1/d}}$, over $d\log(L^{-1/d}/\veps)$ rounds. Using essentially the same argument as that in Section \ref{sec:upperbound}, we can show that the strategy is accurate and private, yielding the query complexity upper bound: 
\begin{align}
N(\veps, \delta, L)\leq &Ld\log(L^{-1/d}/\veps)+dL^{1/d} \nln
=&  dL\log(1/\veps)- L\log L +dL^{1/d}  \nln
\leq& dL\log(1/\veps) -L(\log L-1).  
\label{eq:upperddim}
\end{align}
whenever $\veps<\delta<L^{-1/d}$. The last step follows from the fact that $dL^{1/d} \leq L$ for all $L\geq 2^d$. Note that the above upper bound is similar to that in Theorem \ref{thm:main}, with the addition of a factor $d$ in the first term reflecting the impact of dimensionality. 

We now turn to the query complexity lower bound, by following the same four-step procedure in Section \ref{sec:lowerBound}. There is, however, a crucial difference to be reckoned with: in the original version of the problem, a query can only intersect a single sub-interval in an $s$-uniform partition, a feature that is crucial to the effectiveness of the proportional-sampling  estimator. This is no longer true when $d\geq 2$, because a hyperplane can intersect a large number of sub-cubes in $\calM_s$. That is, a single query could potentially reveal information concerning multiple sub-regions of the unit cube. The strong action-information proximity in the original problem is thus weakened, suggesting that the learner might fare better than in the one-dimensional setting. 

\emph{Steps 1 and 2: Discrete Formulation and Localized Complexity}. The discrete version of the problem is essentially identical to that of the one-dimensional setting, except that we will use the more general definition of the $d$-dimensional $s$-uniform partition, $\calM_s$ (Eq.~\eqref{eq:ddimuniformpart}). Define by $\calV_{s}$ the index set for $\calM_s$: 
\begin{equation}
\calV_{s} \bydef \{(l_1, \ldots, l_d) \in \{1, \ldots, s^{-1}\}^d\}, 
\end{equation}
and by $\calQ^j$ the set of queries whose corresponding hyperplane intersects with the $j$th element of $\calM_\delta$: 
\begin{equation}
\calQ^j \bydef \{Q_i: \exists x\in M_\delta(j) \mbox{ s.t. } \lt<x, Q_i(1:d)\rt>= Q_i(d+1)\}, \quad j \in \calV_s. 
\end{equation}
We have the following localized query complexity lower bound that mirrors Lemma \ref{lem:EQlower}. Note that the only difference is the additional factor of $d$ in Eq.~\eqref{eq:EQj-d}, due to the increase in entropy. 
\begin{lemma}[$d$-Dimensional Localized Query Complexity] 
\label{lem:EQlower-d} 
 Fix $\veps$, $\nu$ and $\delta \in (0,1)$, $\veps<\delta$, and an $(\veps, \nu)$-accurate discrete learner strategy. We have that
\begin{equation}
  \sum_{j\in \calV_{\delta}} \delta \int_{0}^1 \xi_{i,y} \, dy \leq \nu. 
 \label{eq:xitoAvg-d}
\end{equation}
and 
\begin{equation}
\E\lt(|\calQ^j| \bbar J(\delta, X^*) = j, Y=y\rt) \geq (1-\xi_{j,y})d\log(\delta/\veps)- h(\xi_{j,y}), 
 \label{eq:EQj-d}
\end{equation}
for all $j \in \calV_\delta$, and $y\in [0, 1)$. 
\end{lemma}
The proof of Lemma \ref{lem:EQlower-d} is essentially identical to that of Lemma \ref{lem:EQlower}, except for that a sub-interval now becomes a sub-cube, and hence the conditional entropy of the index, $J(\veps, X)$, given the target being in $M_\delta(j)$ increases to $d\log(\delta/\veps)$ from $\log(\delta/\veps)$. 

\emph{Step 3: Proportional Sampling.} We now turn to the analysis of the proportional-sampling estimator, and it is in this portion of the proof that we will see the most significant departure as a result of dimensionality increase. Under a proportional-sampling estimator, the adversary generates $\what{J}^a$ by sampling an index proportional to $|\calQ^j|$, $j\in \calV_\delta$. Crucially, for $j\in \calV$, the equation analogous to Eq.~\eqref{eq:jatoq} is now: 
\begin{align}
\pb(\what{J}^a = J(\delta, X^*) \bbar \calE_{j,y}  ) = 
\E\lt( \frac{|\calQ^j|}{\sum_{j'\in \calV_\delta} |Q^{j'}|}  \Big |\calE_{j,y} \rt). 
\label{eq:JtoP2}
\end{align}
Whereas we had $\sum_{j' \in \calV_\delta} |Q^{j'}| = n$ when $d=1$, this is no longer true when $d\geq 2$,  since the hyperplane associated with a single query can intersect a large number of sub-cubes. The following lemma characterizes the transversality of $(d-1)$-dimensional hyperplanes, showing that $|\calQ|^j$ can be on the order of $\mathcal{O}(\delta^{-(d-1)})$. The proof is based on basic properties of high-dimensional geometry, and is given in Appendix \ref{app:lem:tranversD}. 

\begin{lemma}[Hyperplane Transversality]
\label{lem:tranversD} Fix $d\in \N$. Denote by $H$ a hyperplane in $\R^d$, and denote by $N_H$ the number of sub-cubes in $\calM_\delta$ that intersect $H$. The following is true. 
\begin{enumerate}
\item For all $H$, 
\begin{equation}
N_H \leq g_d\lt(\delta^{-1} + 1\rt)^{d-1}, \quad \delta \in (0,1), 
\end{equation}
where $g_d = {\pi^{\frac{d}{2}}d^{\frac{d+1}{2}} }/{\Gamma\lt(\frac{d+1}{2}\rt)}$ and $\Gamma(\cdot)$ is the gamma function. In other words, there exist positive constants $c_d$ and $\delta_d$, such that   $N_H  \leq c_d \delta^{-(d-1)}$, for all $\delta \in (0, \delta_d)$. 
\item Conversely, there exists an $H$ such that $N_H \geq \delta^{-(d-1)}$. 
\end{enumerate} 
\end{lemma}

Lemma \ref{lem:tranversD} shows that the denominator in Eq.~\eqref{eq:JtoP2} can be on the order of $\mathcal{O}(n\delta^{-(d-1)})$, suggesting that, as the dimension increases, the action-information proximity may drop dramatically and the effectiveness of the proportional-sampling estimator weakened as a result. By combining Eq.~\eqref{eq:JtoP2} and Lemma \ref{lem:tranversD}, we obtain the following analog of Proposition \ref{prop:disc_lowerbound} on the query complexity of the discrete problem: 
\begin{proposition}
\label{prop:disc_lowerbound_ddim}
Let $c_d$ be $\delta_d$ be the positive constants defined  in Lemma \ref{lem:tranversD}. Fix  $\veps$, $\nu$ and $\delta$ in  $ (0, 1)$ and $L\in \N$, such that $\veps < \delta < 1/L <\delta_d$. We have that
\begin{equation}
N^D(\veps, \nu, \delta, L) \geq c_d^{-1}\delta^{d-1}L\lt[ (1-\nu) d\log(\delta/\veps) - h(\nu)\rt], 
\end{equation}
where $h(p)$ is the Shannon entropy of a Bernoulli random variable with mean $p$.
\end{proposition}

\emph{Step 4: Connecting to the Continuous Version.} To derive an analog of Proposition \ref{prop:dis_to_cont}, the only adjustment we need to make is to re-calculate the probability of error when convertings a continuous learner strategy into a discrete one. Specifically, keeping all other portions of the proof of Proposition \ref{prop:dis_to_cont} unchanged, the right-hand side of Eq.~\eqref{eq:probContToDis} will change from $\beta^{-1}$ to $\beta^{-1}d$ via a union bound across all $d$ dimension. We obtain the following analog of Proposition \ref{prop:dis_to_cont}, where the term $\beta^{-1}$ is replaced  $\beta^{-1}d$.
\begin{proposition} 
\label{prop:dis_to_cont_ddim}
Fix $d\in \N$. Fix  $\veps$ and $\delta$ in  $ (0, 1)$ and $L \in \N$, such that $\veps < \delta/(2d)$ and $\delta < 1/L$.   Fix $\beta>0$, such that $\beta\in [2d ,  {\delta}/{\veps}]$. We have that
\begin{equation}
 N (\veps, \delta, L) \geq N^D (\beta\veps, d\beta^{-1}, \delta, L). 
\end{equation}
\end{proposition}
To complete the analysis, define $\gamma = d\beta^{-1} $, and Eq.~\eqref{eq:NoverL} will be modified to become
\begin{align}
\frac{N(\veps, \delta, L) }{L}\cdot \frac{1}{ c_d^{-1}\delta^{d-1}}\geq &   (1-\gamma) d\log\lt(\frac{\delta}{\veps} \gamma\rt) - h(\gamma)\nln
\geq &   (1-\gamma)d \log\lt(\frac{\delta}{\veps} \gamma\rt) + 2(1-\gamma)\log(\gamma) \nln
{\geq} & d\log(\delta/\veps) - \gamma d \log(\delta/\veps) +  (d+2)\log(\gamma), 
\end{align}
whenever $\gamma\leq 1/2$. Suppose that  $\delta/\veps >  2d$. Set $\gamma = {1}/{2}$ or, equivalently, $\beta = 2d$. We have that
\begin{align}
N(\veps, \delta, L) \geq& c_d^{-1}\delta^{d-1} L \lt[\frac{d}{2}\log(\delta/\veps) - (d+2)\rt] \nln
=& c_{d,1} \delta^{d-1} L \log(\delta/\veps) - c_{d,2}L\delta^{d-1}\nln
\geq & c_{d,1} \delta^{d-1} L \log(\delta/\veps) - c_{d,2}L,
\label{eq:NcontLBddim}
\end{align}
where $c_{d,1} = \frac{d}{2} c_d^{-1} $ and $c_{d,2} = c_d^{-1}(d+2)$, and the last inequality follows from the fact that $\delta<1$.  This completes the proof of Theorem \ref{thm:main-ddim}. 

\subsection{Partial Adversary Observation}

We now generalize the model along a different direction starting from the original one-dimensional formulation in Section \ref{sec:intro}. We had initially assumed that the adversary would observe all of the learner's queries exactly, which can be a strong assumption for practical purposes.  In this subsection, we will analyze a generalization of the problem where the adversary is entitled to only partial observations.   Since the power of the adversary is not strengthened under partial observation, any learner strategy and query complexity upper bound developed for the original, stronger adversary model will continue to hold (though there might be room for even better learner strategies), while it is conceivable that the query lower bound will become much weaker under partial observation. Therefore, we will focus on the query complexity lower bound in this subsection.

Formally, we will consider the following general partial observation model. Let $\calU$ be the set of possible observations, and  $f_u: [0, 1)^{n+1} \to  \calU$ an {observation function} which maps the learner's queries $Q\bydef (Q_1,\ldots, Q_n)$ and a random seed, $Y_u$, into a (possibly random) observation, $U=f_u(Q, Y_u)$ in $\calU$. Here, $Y_u$ is a uniform random variable in $[0, 1)$, independent from other parts of the system. The adversary knows $f_u$, but only sees the  observation, $U$, instead of the queries, $(Q_1,\ldots, Q_n)$. 

The function  $f_u$ thus captures any perturbation, noise, or  compression embedded in the adversary's observation. The following definition characterizes the ``quality'' of the observation function in terms of the extent to which the adversary can use $U$ to produce an unbiased estimator for the fraction of queries that lie in $M_\delta(j)$, $|\calQ^j|/n$. 
\begin{definition}[$\zeta$-Biased Observation Function] 
\label{def:zetaUnbiased}
Fix $\delta$ and $\zeta \in (0,1)$. An observation function, $f_u$, is $\zeta$-biased, if under any learner strategy there exist estimators $\{\what{N}^j\}_{j=1,\ldots, 1/\delta}$ which take as input $U=f_u(Q,Y_u)$ and satisfy $\sum_{j = 1}^{1/\delta} \what{N}^j \leq n$, and
\begin{equation}
\frac{\E\lt( \what{N}^j \bbar \calE_{j,y}\rt) }{\E\lt(|\calQ^j| \bbar \calE_{j,y}\rt)} \geq 1-\zeta,
\label{eq:defhatNj}
\end{equation}
for all $j \in \{1,\ldots, 1/\delta\}$ such that $\E\lt(|\calQ^j| \bbar \calE_{j,y} \rt) >0$, where $\calE_{j,y} \bydef \{J(\delta, X^*) = j, Y=y\}$. 
\end{definition} 

We now look at two examples of a partial observation model: 

\emph{Example 1}: In an \emph{additive noise model}, instead of knowing query $Q_i$ exactly, the adversary observes $Q_i+Z_i$, where $Z_i$ is an idiosyncratic Gaussian noise with mean zero  and variance $\sigma^2$. Denote by $\Phi(\cdot)$ the cumulative distribution function of a standard normal distribution. Because the probability density function of the normal distribution is symmetric and monotonically decreasing away from the origin, it is not difficult to show that 
\begin{equation}
\pb(Q_i+Z_i \in M_\delta(j) | Q_i\in M_\delta(j) ) \geq \frac{1}{2} - \Phi(-\delta/\sigma), \quad \forall j = 1, \ldots, 1/\delta, i = 1, \ldots, n. 
\end{equation}
Set the estimator $\what{N}^j$ to be the number of perturbed queries that fall within the $j$th sub-interval of $\calM_\delta$. Clearly, the definition implies that $\sum_{j = 1}^{1/\delta} \what{N}^j = n$, and the above equation implies 
\begin{equation}
\frac{\E\lt( \what{N}^j \bbar \calE_{j,y}  \rt) }{\E\lt(|\calQ^j| \bbar \calE_{j,y}  \rt)} \geq \frac{\E\lt( |\calQ^j| \bbar \calE_{j,y}  \rt) \lt(\frac{1}{2} - \Phi(-\delta/\sigma) \rt)}{\E\lt(|\calQ^j| \bbar \calE_{j,y}  \rt)} = \frac{1}{2} - \Phi(-\delta/\sigma) = 1-\lt( \frac{1}{2} + \Phi(-\delta/\sigma)\rt).
\label{eq:ratioAdditiveNoise}
\end{equation}
Therefore, we conclude that the observation function associated with the additive noise model is $\lt(\frac{1}{2} + \Phi(-\delta/\sigma)\rt)$-biased. Because $\lim_{x\to -\infty}\Phi(x)=0$, in the regime where the magnitude of noise $\sigma$ is substantially smaller than the adversary error tolerance, $\veps$, the observation function becomes essentially $1/2$-biased.  

\emph{Example 2}: In a \emph{random erasure model} with erasure probability $p$, a query $Q_i$ is observed exactly by the adversary with probability $p$, and is not observed at all with probability $1-p$, where the event of an erasure is independent from the rest of the system.  Denote by $N^j$ the number of observed queries that fall within the sub-interval $M_\delta(j)$, and consider the following estimator: 
\begin{equation}
\what{N}^j = \frac{1}{p}N^j\frac{n}{\frac{1}{p}\sum_{j'=1}^{1/\delta} N^{j'}}. 
\label{eq:njcount}
\end{equation}
When $n$ is small or $p$ is large, $\what{N}^j$ can be biased in the sense of Definition \ref{def:zetaUnbiased}. However, if $n$ is sufficiently large or $p$ is sufficiently small, then the denominator in the right-hand side of Eq.~\eqref{eq:njcount} will concentrate around $n$. Since $N^j/p$ is an unbiased estimator for $|\calQ^j|$, it is not difficult to show that in that regime, the ratio ${\E\lt( \what{N}^j \bbar \calE_{j,y}  \rt) }/{\E\lt(|\calQ^j| \bbar \calE_{j,y}  \rt)}$ will approach $1$. Therefore, when $n$ is large or $p$ is small, the observation function in the random erasure model is $\zeta$-biased for a small $\zeta$. 

The following theorem is the main result of this subsection, which generalizes the query complexity lower bound to the partial observation model. 

\begin{theorem}[Query Complexity Lower Bound under Partial Observation] 
\label{thm:main-noisy}
Fix  $\veps$ and $\delta$ in  $ (0, 1)$ and $L \in \N$, such that $\veps < \delta/4$ and $\delta < 1/L$. Suppose that the adversary's observation function is $\zeta$-biased for some $\zeta\in (0,1)$. We have that
\begin{equation}
N(\veps, \delta, L)\geq  (1-\zeta)\lt( L\log(1/\veps) - L\log(2/\delta)- 3L\log\log(\delta/\veps)\rt) .
\end{equation}
\end{theorem}

We may, for instance, apply Theorem \ref{thm:main-noisy} to the additive noise model. Observe that for any $\delta>0$, there exists a $\sigma_0>0$, such that the observation function is $2/3$-biased for all $\sigma \in (0,\sigma_0)$ (cf.~Eq.~\eqref{eq:ratioAdditiveNoise}). Invoking Theorem \ref{thm:main-noisy}, we have that, whenever $\sigma \in (0,\sigma_0)$, 
\begin{equation}
N(\veps, \delta, L)\geq  \frac{1}{3}\lt( L\log(1/\veps) - L\log(2/\delta)- 3L\log\log(\delta/\veps)\rt), 
\end{equation}

To prove Theorem \ref{thm:main-noisy}, we need only modify  the proportional-sampling estimator of Section \ref{sec:proportionalSample}. Fix a learner strategy in $\Phi_n$, and let $\what{N}^j$ be the estimators in Definition \ref{def:zetaUnbiased}. Consider the proportional sampling estimator in Definition \ref{def:psample}, but rather than sampling an index $j$ with a probability proportional to $|\calQ_j|$, let the index instead be sampled with a probability proportional to  $\what{N}^j$. With this mortification, we have that 
\begin{align}
\pb\lt( \what{J}^a =  J(\delta, X^*) \bbar  \calE_{j,y} \rt)  
=  \E\lt( \lt. \frac{\what{N}^j}{\sum_{j'=1}^{1/\delta} \what{N}^{j'}} \, \,  \rt| \,\, \calE_{j,y} \rt) 
\sk{a}{\geq}  (1-\zeta)\frac{1}{n} \E\lt( |\calQ^j| \bbar  \calE_{j,y}  \rt),
\label{eq:jatoq-noisy}
\end{align}
where step $(a)$ follows from the fact that the observation function is $\zeta$-biased. Comparing Eq.~\eqref{eq:jatoq-noisy} to the original Eq.~\eqref{eq:jatoq}, we notice that the only difference lies in the additional constant factor $(1-\zeta)$. It is not difficult verify that the rest of the proof the original lower bound can be carried out in essentially an identical manner.

\section{Concluding Remarks}
\label{sec:conclusion}

The main result of the present paper is a tight query complexity lower bound for the Bayesian Private Learning problem, which, together with an upper bound in \citet{tsixuxu2017}, shows that the learner's query complexity depends multiplicatively on the level of privacy, $L$:  if an $\veps$-accurate learner wishes to ensure that an adversary's probability of making a $\delta$-accurate estimation  is at most $1/L$, then she needs to employ on the order of $L\log(\delta/\veps)$ queries. Moreover, we show that the multiplicative dependence on $L$  holds even under the more general models of high-dimensional queries and partial adversary observation. To prove the lower bound, we develop a set of  information-theoretic arguments which involve, as a main ingredient, the analysis of proportional-sampling adversary estimators that exploit the  action-information proximity inherent in the learning problem. The present work leaves open a few interesting directions: 

The current upper and lower bounds are not tight in the regime where the adversary's error criterion, $\delta$, is significantly smaller than $1/L$. Making progress in this regime is likely to require a more delicate argument and possibly new tools.   

While we are able to extend the proof to the more general models in Section \ref{sec:generalizations}, the query complexity bounds are no longer tight, partially because the proportional-sampling estimators are not as effective under weakened action-information proximity. It would be interesting to see whether the gap between the upper and lower bounds can be shrunk by constructing more sophisticated learner strategy and adversary estimators. For instance, in the high-dimensional linear query model,  while a query may intersect multiple sub-cubes, the sub-cube that contains the target is likely to be associated with more concentrated intersections than those that do not. This seems to suggest that the adversary might benefit by weighing the ``susceptibility'' of the sub-cubes \emph{non-linearly} as a function of the queries' empirical densities, instead of the linear weighing prescribed by the proportional-sampling estimators. 

Our query model assumes that the responses to the learner's queries are noiseless, and it will be interesting to explore how may the presence of noisy responses \citep{rivest1980coping, ben2008bayesian, waeber2013bisection} impact the design of private query strategies. For instance, a natural generalization of the bisection search algorithm to the noisy setting is the Probabilistic Bisection Algorithm \citep{horstein1963sequential, waeber2013bisection}, where the $n$th query point is the median of the target's posterior distribution in the $n$th time slot. It is conceivable that one may construct a probabilistic query strategy  analogous to the Replicated Bisection strategy by replicating queries in $L$ pre-determined sub-intervals. However, it appears challenging to prove that such replications preserve privacy. Since noisy responses weaken the power of the learner,  the query complexity lower bound that we have developed in Theorem \ref{thm:main} immediately extends to the noisy setting; it is however unclear how this lower bound can be improved to {match} that of a private query strategy.

One may also want to extend our analysis to more complex models of  active learning or online (convex) optimization. For instance, since the current model corresponds to actively learning an threshold function via function-value queries (see Remark \ref{remk:activelearn}), it would be interesting to know how to design private strategies for learning richer functional classes (e.g., smooth convex functions).  Alternatively, one may also consider a model in which each query reveals to the learner the full gradient of a function at the queried location, instead of only the sign of the gradient as in the present model. 

 Finally, our present discussion of the action-information proximity is rather qualitative, and it would be desirable to formulate a precise and general framework to articulate the action-information proximity in sequential learning problems, and better understand how it relates to the cost of privacy.

\bibliographystyle{apalike}
\bibliography{bibliography/bayesprivate.bib}

\ifx \useplain\undefined
\begin{APPENDIX}{}
\else
\appendix
\fi

\normalsize

\section{Proofs}
\subsection{Proof of Proposition \ref{prop:upper}} 
\label{app:prop:upper}
\emph{Proof.} Denote by $\tilde{Q}_l$ the query submitted in the $l$th sub-interval during the last round of Phase 2 of the Replicated Bisection strategy. Note that since the positions of the queries relative to their respective sub-interval are identical in each round, we must have that
\begin{equation}
|\tilde{Q}_l -\tilde{Q}_{l'} | \geq \frac{1}{L}, \quad \forall l, l'\subset \{1,\ldots, L\}, l\neq l'. 
\label{eq:QtilApart}
\end{equation}
By the end of the second phase, the adversary knows that the target belongs to the sub-interval $[\tilde{Q}_l-\veps, \tilde{Q}_l+\veps)$ for some $l\in \{1,\ldots,L\}$, but not more than that. Formally, it is not difficult to show that, almost surely, the posterior density of $X^*$ is
\begin{equation}
f_{X^*}(x|(Q_1,\ldots, Q_n)) = \frac{1}{2L\veps}, \quad \forall x \in \cup_{l=1}^L [\tilde{Q}_l-\veps, \tilde{Q}+\veps), 
\label{eq:XdensityonQ}
\end{equation}
and $f_{X^*}(x|(Q_1,\ldots, Q_n))=0$ everywhere else. Recall that  $\veps<\delta$  and $\delta<1/L$ by assumption, we have that 
\begin{equation}
\delta/2<-\veps+ 1/L,
\end{equation}
where the right-hand side corresponds to the distance between two adjacent intervals $[\tilde{Q}_l-\veps, \tilde{Q}+\veps)$. 
In light of Eq.~\eqref{eq:QtilApart}, this implies that for any interval $G \subset [0, 1)$ with length $\delta$, 
\begin{equation}
\mu^{\calL}\lt( G \cap \lt(\cup_{l=1}^L [\tilde{Q}_l-\veps, \tilde{Q}+\veps) \rt)\rt) \leq 2\veps,\quad \mbox{almost surely,}
\end{equation}
where $\mu^\calL(\cdot)$ is the Lebesgue measure.  Combining the above inequality with Eq.~\eqref{eq:XdensityonQ}, we conclude that, for any adversary estimator $\what{X}^a$, generated based on $Q$, we have 
\begin{equation}
\pb\lt(|\what{X}^a -  X^*| \leq \delta/2 \bbar (Q_1,\ldots, Q_n) \rt) \leq 1- \frac{L-1}{L} = \frac{1}{L}, \quad \mbox{almost surely}.
\end{equation}
This shows that the Replicated Bisection strategy is $(\delta,L)$-private. 
\qed

\subsection{Proof of Lemma \ref{lem:tranversD}}
\label{app:lem:tranversD}

\emph{Proof.} The lower bound on $N_H$ follows from observing that the hyperplane $H = \{(x_1,\ldots, x_d) \in \R^d: x_1 = 0.5\}$, we have $N_H= \delta^{-(d-1)}$. We now prove the upper bound. Denote by $ \calJ^H$ the indices of the sub-cubes in $\calM_\delta$ that intersect $H$, and by $\tilde H$ the cross section of the unit cube along $H$: 
\begin{equation}
\tilde H = H \cap [0, 1)^d. 
\end{equation}
Note that the diameter of a length-$\delta$ cube in $d$ dimensions is equal to  $\delta\sqrt{d}$. Denote by $B_d(x,r)$ a $d$-dimensional ball centered at $x$ with radius $r$. It thus follows that 
\begin{equation}
\cup_{j \in \calJ^H}M_\delta(j) \subset (B_d(0,\delta\sqrt{d}) + \tilde{H}) \bydef \tilde{H}_B, 
\end{equation}
where the addition on the right-hand side denotes the Minkowski sum. This further implies that 
\begin{equation}
\vol_d \lt(\cup_{j \in \calJ^H}M_\delta(j) \rt) \leq \vol_d(\tilde{H}_B),  
\end{equation}
where $\vol_d(\cdot)$ represents the $d$-dimensional volume. Because the diameter of $[0, 1)^d$ is $\sqrt{d}$, the cross section $\tilde{H}$ is contained by some $(d-1)$-ball with radius $\sqrt{d}/2$. It is not difficult to show that
\begin{align}
 \vol_d(\tilde{H}_B) =& \vol_d( B_d(0,\delta\sqrt{d}) + \tilde{H} ) \nln
 \leq& \vol_{d-1}\lt(B_{d-1} \lt( 0,\delta\sqrt{d}+ {\sqrt{d}}/{2} \rt) \rt)  \delta\sqrt{d} \nln
= & \frac{\pi^{\frac{d-1}{2}}}{\Gamma\lt(\frac{d-1}{2}+1\rt)}\lt(\delta\sqrt{d}+ {\sqrt{d}}/{2}\rt)^{d-1} \delta\sqrt{d} 
\end{align}
where the inequality follows from the fact that $B_d(0,\delta\sqrt{d}) + \tilde{H}$ is contained in a cylinder with radius $\delta\sqrt{d}+ {\sqrt{d}}/{2}$ and height $\delta\sqrt{d}$. Because $\cup_{j \in \calJ^H}M_\delta(j)$ consists of $N_H$ disjoint cubes with volume $\delta^d$ each, we have that
\begin{align}
 N_H = & \frac{\vol_d(\cup_{j \in \calJ^H}M_\delta(j))}{\delta^d}
 \leq  \frac{ \vol_d(\tilde{H}_B)}{\delta^d}  \nln
 \leq & \frac{\pi^{\frac{d-1}{2}}}{\Gamma\lt(\frac{d-1}{2}+1\rt)}\lt(\delta\sqrt{d}+ {\sqrt{d}}/{2}\rt)^{d-1} \sqrt{d} \delta^{d-1} \nln
 \leq &\frac{\pi^{\frac{d}{2}}d^{\frac{d+1}{2}} }{\Gamma\lt(\frac{d+1}{2}\rt)}\lt(\delta^{-1} + 1\rt)^{d-1}. 
 \end{align}
This proves the upper bound on $N_H$. \qed

\section{Replicated Bisection Search is not Differentially Private}
\label{sec:BPSnotDP}

We highlight in this section the difference between the privacy framework  considered in this paper (goal-oriented) and that of differential privacy (universal). We will demonstrate that the Replicated Bisection strategy (Section \ref{sec:upperbound}), while  private according to our formulation, is \emph{never} differentially private. The discussion in this section is not meant to be very  formal, but we will aim to elucidate the main idea in a clear and analytical manner. Furthermore, there may be multiple ways to cast our problem into the differential privacy framework, and we will choose one that appears most natural. 

We begin with a brief review of the basic concepts and terminology in differential privacy (cf.~Chapter 2 of \citet{dwork2014algorithmic}). Let $\calU$ be an arbitrary set, and an element $u \in \calU$ be called a \emph{(data) entry}. Fix $n\in \N$. A vector consisting of data entries, $\bar{u} = (\bar u_1, \bar u_2, \ldots, \bar u_n) \in \calU^n$, is called a \emph{database}. For two databases $\bar u$ and $\bar u'$ in $\calU^n$, denote by $d_1(\bar u, \bar u')$ the number of data entries that differ between them: $d_1(\bar u, \bar u') = \sum_{i=1}^n \mathbb{I}(\bar u_i \neq \bar u')$.  Let $\calY$ be a set of \emph{observations}. Finally, a \emph{mechanism}, $M : \calU^n \to \calY$, is a (randomized) mapping from the space of databases to that of  observations.  

Roughly speaking, differential privacy demands that the observation $M(\bar u)$ should not reveal  the nature of \emph{individual entries} in the database, $\bar{u}$. Specifically, if one were to perturb one entry in a database, $\bar u$, to create a second database, $\bar u'$, then the distributions of $M(\bar u)$ and $M(\bar u')$ should be ``close'' under an appropriate metric. Formally, we have the following definition (e.g., Definition 2.4, \citet{dwork2014algorithmic}): 
\begin{definition} Fix $n\in \N$, $\delta \in [0,1)$  and  $\veps \in \rp$. A mechanism $M$ is $(\veps, \delta)$-differentially private if for all $\calS \subset \calY$, and $\bar u$ and $\bar u'$ in $\calU^n$ such that $d_1(\bar u, \bar u')=1$, we have
\begin{equation}
\pb(M(\bar u) \in \calS) 	\leq \exp(\veps)\pb(M(\bar u') \in  \calS) + \delta, 
\end{equation}
where the probability is measured with respect to the randomization in $M$. 
\end{definition}
For instance, a $(0,0)$-differentially private mechanism will produce an identical output distribution under any input database. More generally, a smaller value of $\veps$ and $\delta$ means that the mechanism is more private. 

We now cast our problem in the above framework. 
\begin{enumerate}
\item \emph{Database}: since the information that the learner wishes to conceal is the target, $X^*$, a natural way to encode would be to express $X^*$ in its binary expansion up to $\log(1/\veps)$ bits of precision (beyond which the learner will have no need to discern). Let $n= \log(1/\veps)$, and let $ \sum_{i=1}^n \bar u_ i 2^{-i} $ be the $n$-bit binary expansion of $X^*$. The database of interest will therefore be $\bar u = (\bar u_1, \ldots, \bar u_n)$. We have $\calU = \{0,1\}$ and $\calU^n = \{0,1\}^n$. 
\item \emph{Mechanism and Observations}: the mechanism, $M$, in our case would be the learner's strategy. That is, $M$ takes as input the database, $\bar u$, and produces as an \emph{observation} the sequence of queries, $(Q_i)_{i =1, \ldots, N}$. We  have $\calY = [0,1)^N$. 
\end{enumerate}

We now examine how the criterion of differential privacy relates to our notion of privacy (Definition \ref{def:dLprivacy}). In one direction, it is clear that there is a differentially private mechanism (with sufficiently small $\veps$ and $\delta$) that functions as a $(\delta,L)$-private query strategy: simply look at the case where $\veps= \delta=0$, in which case the strategy would lead to the same distribution of queries regardless of what the true target is! 

It is in the other direction that we see that differential privacy  can be substantially more restrictive. Let the mechanism $M$ be associated with the Replicated Bisection strategy (Section \ref{sec:upperbound}). For concreteness, consider the case of $L=2$, and the learner strategy would conduct two identical copies of bisection search in the sub-intervals $[0,1/2)$ and $(1/2,1)$. Under this strategy, the adversary will know the exact {relative} location of the target (up to $\veps$ error) with respect to each sub interval, but does not know {which} of the two sub-interval contains the target. In other words, $M$ reveals {all but the first} bit of $\bar u$, which encodes the identity of the ``true'' sub-interval!\footnote{Technically, the adversary also does not know the very last bit of $\bar u$. But this detail does not alter our subsequent argument in any significant way, since all bits between the first and the last will be known. }

From the above observation, we can deduce that the Replicated Bisection strategy can never be differentially private: let $\bar u $ be the all-zero vector, and $\bar u'$ a vector where all but the {second} bit is zero. That is,  $\bar u$ and $\bar u'$ differ in the second bit, and $d_1(\bar u, \bar u')=1$.  Let $\calS_0$ be the smallest subset of $\calY$ such that $\pb(M(\bar u) \in \calS_0) = 1.$ Intuitively, $\calS_0$ is the set of all ``plausible'' observations under $\bar u$. Since the mechanism always reveals the second bit of $\bar u$, it follows that any observation that's ``plausible'' under $\bar u$ must be ``implausible'' under $\bar u '$. That is, 
\begin{equation}
\pb(M(\bar u') \in \calS_0) = 0. 
\end{equation}
This shows that $M$ is {not} $(\veps, \delta)$-differentially private for any non-trivial choice of $\veps$ and $\delta$ ($\veps >0$ and $\delta \in (0,1)$). The same analysis extends easily to $L\geq 3$, since, in general, the mechanism associated with the Replicated Bisection strategy reveals all but the first $\log L$ bits of the database. 

More importantly, the above example also shows \emph{why} differential privacy is more restrictive. In differential privacy, the mechanism aims to conceal  {any} (one-entry) perturbation of the database. In contrast, the Replicated Bisection strategy has no trouble giving away the less significant bits of the database, and is only concerned with obfuscating perturbations in the leading bits of $\bar u$ which would result in ``large'' changes in the target's location as measured by the $L_1$ distance. In other words,  a goal-oriented notion of privacy naturally induces  {different} degrees of ``importance''  across data entries (bits), a key feature that is not captured by differential privacy, who treats all data entries with {equal} importance. 

\ifx \useplain\undefined
\end{APPENDIX}
\fi

\end{document}